\documentclass{article}

\usepackage[utf8]{inputenc}
\usepackage[T1]{fontenc}
\usepackage[nonatbib,preprint]{neurips_2025}

\usepackage[
  backend=biber,
  style=numeric,
  maxnames=10,
  minnames=1,
  sorting=nyt,
  url=true,
  doi=true,
  eprint=false
]{biblatex}
\renewbibmacro*{url+urldate}{%
  \printfield{url}%
}

\usepackage[utf8]{inputenc} 
\usepackage[T1]{fontenc}    
\usepackage{hyperref}       
\usepackage{url}            
\usepackage{booktabs}       
\usepackage{amsfonts}       
\usepackage{microtype}      
\usepackage{xcolor}         
\usepackage{graphicx}
\usepackage{amsmath}
\usepackage{subcaption}
\usepackage{tcolorbox}
\usepackage{longtable}
\usepackage{makecell}
\usepackage{float}

\title{AgentCaster: Reasoning-Guided Tornado Forecasting}

\author{%
  Michael Chen\\
  Department of Computing + Mathematical Sciences\\
  California Institute of Technology\\
  Pasadena, CA\\
  \texttt{mhchen@caltech.edu}\\
}

\begin{document}

\maketitle

\begin{abstract}
  There is a growing need to evaluate Large Language Models (LLMs) on complex, high-impact, real-world tasks to assess their true readiness as reasoning agents. To address this gap, we introduce AgentCaster, a contamination-free framework employing multimodal LLMs end-to-end for the challenging, long-horizon task of tornado forecasting. Within AgentCaster, models interpret heterogeneous spatiotemporal data from a high-resolution convection-allowing forecast archive. We assess model performance over a 40-day period featuring diverse historical data, spanning several major tornado outbreaks and including over 500 tornado reports. Each day, models query interactively from a pool of 3,625 forecast maps and 40,125 forecast soundings for a forecast horizon of 12-36 hours. Probabilistic tornado-risk polygon predictions are verified against ground truths derived from geometric comparisons across disjoint risk bands in projected coordinate space. To quantify accuracy, we propose domain-specific TornadoBench and TornadoHallucination metrics, with TornadoBench highly challenging for both LLMs and domain expert human forecasters. Notably, human experts significantly outperform state‑of‑the‑art models, which demonstrate a strong tendency to hallucinate and overpredict risk intensity, struggle with precise geographic placement, and exhibit poor spatiotemporal reasoning in complex, dynamically evolving systems. AgentCaster aims to advance research on improving LLM agents for challenging reasoning tasks in critical domains.
\end{abstract}

\section{Introduction}
\label{intro}

LLMs have rapidly progressed from text-only pattern recognizers to general-purpose reasoning agents capable of planning, using tools, and operating in multi-turn interactions \cite{brown_language_2020, chang_survey_2024, naveed_comprehensive_2024, yi_survey_2024, ma_agentboard_2024, wang_mint_2024}. As these models are increasingly envisioned for autonomous roles, evaluating their true capabilities on more challenging and higher impact problems becomes paramount \cite{wang_survey_2024}. Current benchmarks often fall short. Many focus on relative performance between models rather than absolute capability on real-world tasks, suffer from data contamination, or lack the complexity to probe sophisticated reasoning abilities in real-world contexts \cite{white_livebench_2025, phan_humanitys_2025}. This evaluation gap inhibits our understanding of both LLM limitations and progress, particularly in domains where reliable performance is critical.

Severe convective weather represents precisely such a domain. Predicting tornadoes carries immense importance; from 2010 through 2024, tornadoes in the United States caused over USD 25 billion in property damage and claimed more than 1,200 lives \cite{noauthor_storm_nodate}. Human forecasters at the NWS Storm Prediction Center (SPC) must synthesize heterogeneous high-resolution numerical weather prediction (NWP) fields, examine vertical atmospheric profiles, reason across extensive geographic areas and timeframes, and ultimately produce nested probabilistic polygons that communicate risk to emergency managers and the public \cite{corfidi_birth_1999}. However, despite decades of research, tornado forecasting remains notoriously challenging.

Tornado forecasting is a strong evaluation task for LLM agents. It represents a uniquely challenging reasoning task for AI agents, requiring the synthesis and interpretation of vast, heterogeneous, spatiotemporal meteorological data under varying uncertainty. The task demands integration of visual map data with point-based atmospheric profiles (soundings) and translate this understanding into precise, actionable geographic predictions. While machine learning has made strides in large-scale weather prediction \cite{lam_learning_2023, pathak_fourcastnet_2022}, evaluating the agentic reasoning capabilities of LLMs in this interactive forecasting process remains unexplored.

\begin{figure}
    \centering
    \includegraphics[width=0.99\linewidth]{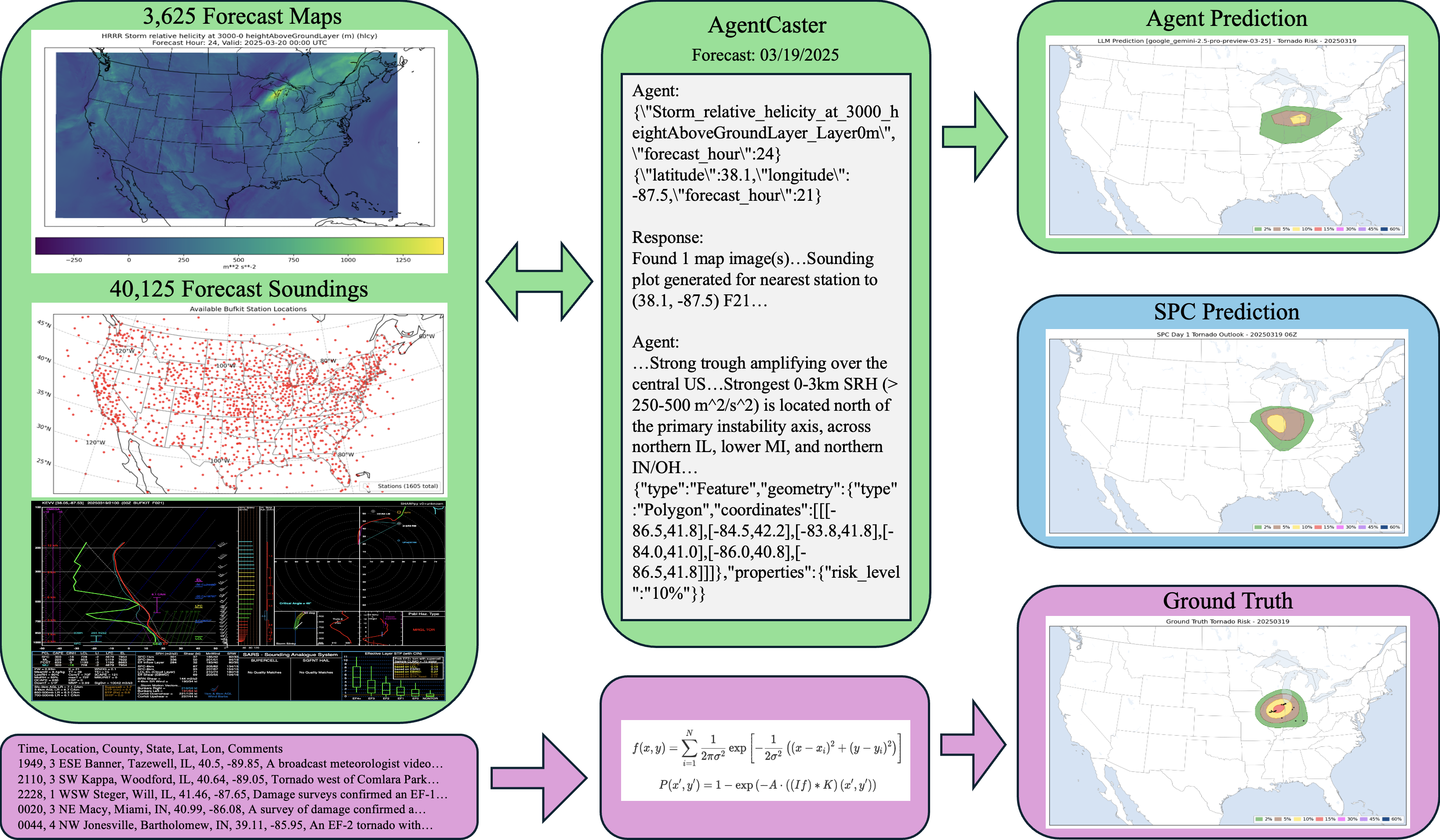}
    \caption{A simplified overview of the AgentCaster framework. LLM agents act as AI meteorologists by first requesting and analyzing forecast maps, then passing specific longitudes and latitudes which are processed to return targeted atmospheric soundings. Agents reason about severe weather dynamics and when confident, generate probabilistic tornado risk predictions as geospatial polygons. These predictions are evaluated against ground truths derived from observed tornado reports through practically perfect forecasts \cite{hitchens_objective_2013} and compared with domain expert SPC forecast baselines.}
    \label{fig:agentcaster_overview}
\end{figure}

To address this evaluation gap, we develop a framework that can rigorously test LLM capabilities in a real-world forecasting environment. We introduce AgentCaster, a novel, contamination-free evaluation framework that assesses multimodal LLM agents end-to-end on tornado forecasting. As shown in Figure \ref{fig:agentcaster_overview}, within AgentCaster, LLMs function as AI meteorologists, interactively querying a rich archive of historical, high-resolution weather forecast data. Mimicking human forecaster workflows, agents first request and analyze relevant forecast maps (e.g., convective inhibition, convective instability) from a pool of 145 available product types, with each product available hourly across the 12-36 hour forecast period. Based on map assessments, they can request specific forecast soundings at geographic coordinates of interest, enabling detailed examination of vertical profiles while operating under daily quotas that encourage strategic resource allocation. Finally, agents synthesize their findings to produce probabilistic tornado risk predictions as geospatial polygons in standard GeoJSON format, analogous to official SPC outlooks.

Evaluating these complex, spatially precise outputs necessitates domain-specific metrics. We propose TornadoBench, an evaluation score based on direct geometric comparisons between the agent's predicted risk polygons and a ground truth derived from observed tornado reports. TornadoBench calculates risk-weighted Intersection over Union (IoU) across disjoint risk bands in projected coordinate space, rewarding accurate placement, extent, and nesting of risk areas. To quantify critical failure modes, we introduce TornadoHallucination metrics (Simple and Hard) that measure false alarm frequency and severity, penalizing predictions of risk on days where less than the minimum 2\% risk occurred or complete misplacement of risk areas on risk days. Our evaluation spans 40 days of diverse weather conditions, including major tornado outbreaks and over 500 tornado reports.

Our contributions include: (1) AgentCaster, a multimodal, interactive, and contamination-free agent framework for evaluating LLM reasoning on the challenging and real-world task of tornado forecasting using daily generated high-resolution forecast data; (2) domain-specific evaluation metrics based on geometric verification against ground truths; (3) a curated 40-day benchmark dataset comprising 145,000 processed forecast maps, on-demand generation for 1,605,000 forecast soundings, SPC outlooks for baseline comparison, and processed ground truth tornado reports; (4) initial evaluation of state-of-the-art multimodal LLMs against human expert baselines; and (5) release of all code and datasets to facilitate reproducibility and further research.

We hope AgentCaster will catalyze research on \textit{high-impact, real-world reasoning tasks} and motivate progress towards agents that can meaningfully assist human experts in critical domains.

\section{Related Work}
\label{related}

\paragraph{Benchmarking LLMs and agents.} Recent years have seen rapid development in benchmarks to keep pace with large language models \cite{brown_language_2020}, with increasingly complex reasoning assessments \cite{hendrycks_measuring_2021, wang_mmlu-pro_2024, srivastava_beyond_2023, liang_holistic_2023, chollet_arc_2025}. However, many existing benchmarks are facing saturation, with state-of-the-art models approaching or exceeding human-level performance. Some works \cite{white_livebench_2025} attempt to address contamination by using updated information sources, while others \cite{phan_humanitys_2025} position themselves as testing at the frontier of human knowledge. The emergence of agent frameworks has introduced new benchmarking challenges. A few approaches \cite{liu_agentbench_2023, yang_swe-agent_2024} evaluate LLMs across diverse environments, and others \cite{zhou_webarena_2024, qin_toolllm_2023, ruan_identifying_2024} assess tool use in various environments.

\paragraph{Multimodal reasoning.} The rise of Vision-Language Models (VLMs) \cite{li_survey_2025, zhang_vision-language_2024, radford_learning_2021, zhu_minigpt-4_2023, yang_dawn_2023} has spurred new approaches to evaluating visual-language integration \cite{chen_are_2024}. Benchmarks for spatial reasoning \cite{wang_is_2024, li_topviewrs_2024} reveals that multimodal models struggle with spatial relationships, often performing worse than text-only LLMs on spatial tasks given preference between visual and textual context. Some temporal reasoning benchmarks \cite{su_living_2024} have also been explored. Others focus on spatiotemporal understanding through videos \cite{chen_rextime_2024} or egocentric spatiotemporal reasoning \cite{wu_st-think_2025}; in general, evaluations show that models struggle to track changes over time, integrate spatiotemporal information, and understand causality.

\paragraph{Expert domain tasks.} Specialized knowledge domains increasingly serve as benchmarks for LLMs, with notable examples in medicine \cite{jin_what_2020, singhal_large_2022, yao_medqa-cs_2024}, law \cite{guha_legalbench_2023}, and finance \cite{xie_finben_2024}; furthermore, domain-specific evaluations can highlight gaps between knowledge retrieval and the nuanced reasoning required for expert-level tasks. These evaluations offer several advantages: they require deep expertise, can integrate multiple reasoning modes, and feature well-defined evaluation criteria with established human expert performance. A common limitation is that such benchmarks rely on static question-answering or classification based on domain corpora. In contrast, AgentCaster utilizes tornado science as an expert domain but evaluates a dynamic, interactive problem-solving forecasting process.

\paragraph{Machine learning for weather forecasting.} Weather forecasting has had significant advances through deep learning approaches. Previous work with global models \cite{lam_graphcast_2023, pathak_fourcastnet_2022, bi_pangu-weather_2022} have demonstrated competitive performance with traditional NWP methods. Some experimental systems \cite{heinselman_warn--forecast_2023} update convection-allowing ensembles frequently to extend warning lead times. However, these approaches typically operate directly on gridded NWP data, maintaining a closed-loop architecture that differs fundamentally from the human forecasting process \cite{lam_graphcast_2023, pathak_fourcastnet_2022, bi_pangu-weather_2022, heinselman_warn--forecast_2023, hill_forecasting_2020}. Tornado nowcasting has been explored with CNNs with some success \cite{lagerquist_deep_2020, veillette_benchmark_2024}, but nowcasting is an entirely different process from forecasting \cite{church_tornado_1993}. AgentCaster is the first framework to deploy machine learning for weather forecasting through an interactive, human-like workflow.

\section{AgentCaster}
\label{AgentCaster}

\subsection{Framework Overview}

AgentCaster is an interactive environment where an LLM agent is placed in the role of an AI meteorologist tasked with issuing a tornado risk forecast for the Continental United States (CONUS). Agents make sequential requests for meteorological data products using a defined set of tools. They begin with access to a wide array of forecast maps and can subsequently request vertical atmospheric profiles for specific locations and times. The agent must predict the probability of a tornado occurring within 25 miles of any point during a 24-hour period from 12:00 UTC on the target date to 12:00 UTC the following day, aligning with operational forecasting timelines used by human meteorologists. For all experiments reported here, we freeze a contiguous 40-day benchmark window (March 1, 2025 to April 9, 2025) to ensure fair composition and reproducibility, even though the framework is designed for live daily forecasting.

AgentCaster's design enables: (1) \textit{realistic assessment of domain expertise} by requiring reasoning similar to expert human forecasters; (2) \textit{interactive exploration} through deliberate tool usage to analyze heterogeneous data; and (3) \textit{contamination-free evaluation} using rolling numerical weather prediction archives. Distinct from text-based or purely simulated environments, AgentCaster dynamically integrates real-world, multimodal meteorological data (including on-demand visual sounding generation triggered by agent requests) within an interactive loop, as illustrated by Figure \ref{fig:agentcaster_overview}. AgentCaster is also \textit{extensible}, allowing for the future inclusion and modification of different NWP models, prediction objectives, or prediction horizons.

\subsection{Meteorological Data Sources}

AgentCaster utilizes archived data from daily runs of the High-Resolution Rapid Refresh (HRRRv4) \cite{dowell_high-resolution_2022} model, processed into formats suitable for multimodal LLM inputs. The HRRRv4 is the state-of-the-art, 3-km resolution, convection-allowing numerical weather prediction system operated by NOAA, built on the WRF-ARW dynamical core \cite{powers_weather_2017}.

For each day, we process the 00:00 UTC HRRR model run to extract and visualize all 145 available map products. These include convective parameters (CAPE, CIN), wind fields (shear, helicity), moisture variables, temperature profiles, and simulated radar reflectivity, among others. Each variable is available for all forecast hours within the prediction window (12-36), resulting in 3,625 distinct map images per day. A data processing pipeline parses this gridded data from raw GRIB2 files and generates these visualizations, rendered onto a consistent map projection covering the CONUS and overlaid with geographic references. See Appendix \ref{app:dataset_details} for the full list of forecast map products.

To access full vertical atmospheric structure near any given point, the framework provides forecast soundings derived from HRRR BUFKIT data. These are generated \emph{on-demand} during the agent's interaction. When an agent requests a sounding for a specific latitude, longitude, and forecast hour, the system identifies the nearest available forecast point from the BUFKIT dataset (from a pool of 1,605 stations available each hour as displayed in Figure \ref{fig:agentcaster_overview}) via computation of Haversine distance. The vertical profile data for that location and time is then extracted and rendered as a standard skew-T log-P diagram using a modified SHARPpy program \cite{blumberg_sharppy_2017}. This visualization includes temperature and dew point profiles, wind barbs, and calculated thermodynamic and kinematic parameters. This on-demand generation, coupled with a daily quota (defaulting to 50 requests), encourages: (1) \textit{targeted geographic focus}; (2) \textit{efficient context window use}; and (3) \textit{strategic decision-making} under resource constraints.

\subsection{Agent Interaction Loop}

The agent utilizes the meteorological data sources within a multi-turn conversational loop designed to mimic an iterative analysis and forecasting process. The loop proceeds through several phases. Complete prompts and code are given in Appendix \ref{app:prompts_code}.

The loop begins with the agent receiving the initial prompt defining the task, date, and available tools. The agent typically starts by calling the \texttt{list\_available\_map\_types} tool to understand the scope of map data for the day. Based on this or subsequent analysis, the agent actively queries the environment by invoking the \texttt{request\_hrrr\_map} tool or the \texttt{request\_sounding} tool.

The AgentCaster backend processes the agent's request. For maps, it retrieves the corresponding pre-generated file. For soundings, it executes the on-demand generation pipeline. The system then responds to the agent with a message containing confirmation text and, if successful, the requested image(s) embedded directly within the message structure. Sounding responses also include the remaining daily quota. If a request fails (e.g., map not found, quota exceeded, sounding generation error), an informative error message is returned instead.

The agent's analysis of the received multimodal information drives the next step. It may identify areas of interest on a map and request specific soundings within those areas using \texttt{request\_sounding} to examine vertical details, respecting the daily quota. Alternatively, it might request different map types or forecast hours via \texttt{request\_hrrr\_map} to build a more comprehensive spatiotemporal understanding. This iterative cycle of request, receive, and analyze continues until the agent deems its analysis sufficient.

Once confident in its assessment, the agent concludes the interaction by invoking the \texttt{submit\_tornado\_prediction} tool. This requires providing the final forecast as a single, structured GeoJSON \texttt{FeatureCollection} string within the \texttt{prediction\_geojson} argument. This GeoJSON must adhere to specific formatting rules, defining distinct polygonal areas for each standard SPC tornado risk category (2\% to 60\%) and ensuring correct spatial nesting (higher risks contained within lower risks). Upon invocation of this tool, the interaction for that forecast day is complete.

\section{TornadoBench and TornadoHallucination}
\label{BenchHallu}

\subsection{Ground Truth Generation}

\begin{figure}[h!]
    \centering
    \begin{minipage}{0.49\linewidth}
        \centering
        \includegraphics[width=\linewidth]{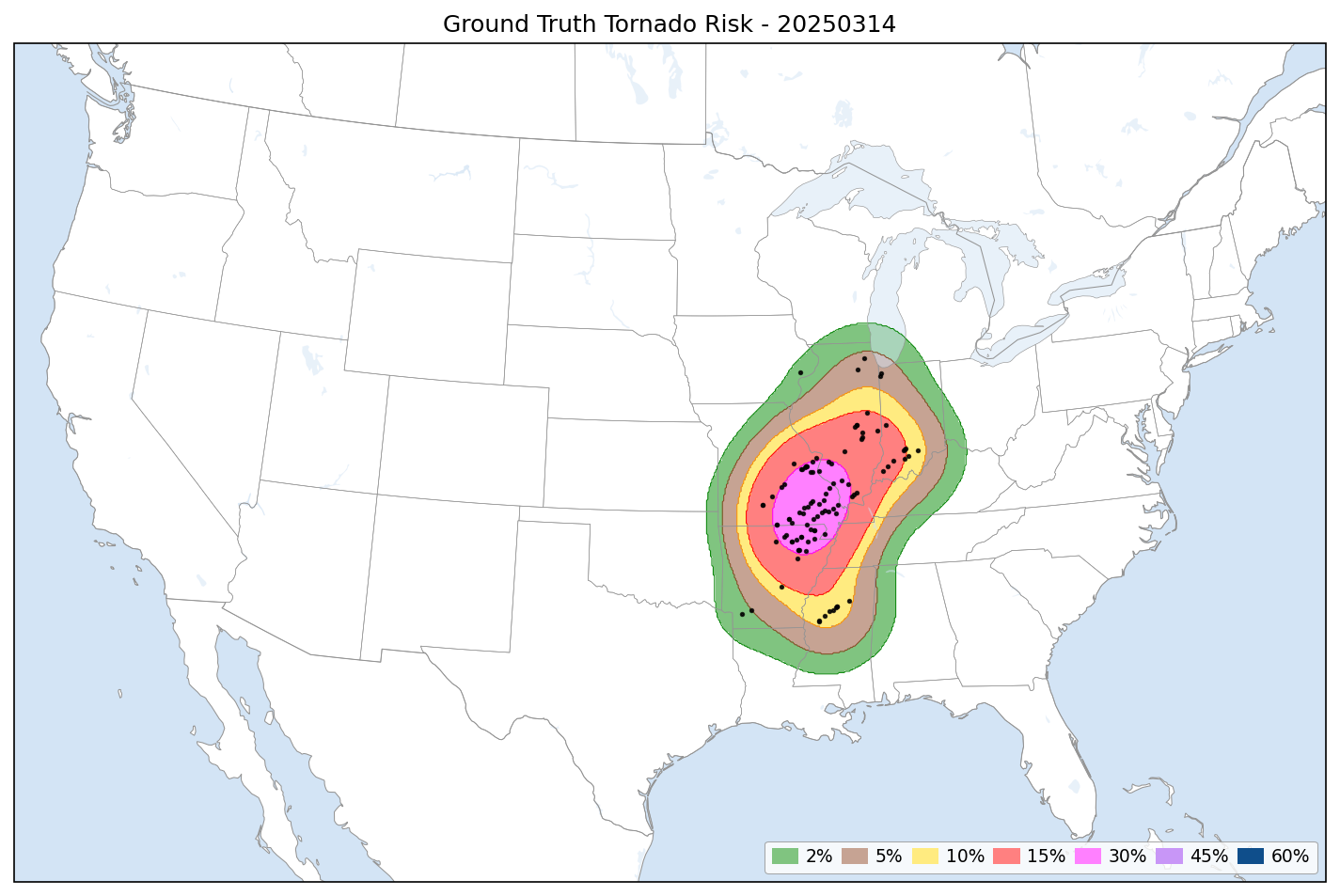}
    \end{minipage}\hfill
    \begin{minipage}{0.49\linewidth}
        \centering
        \includegraphics[width=\linewidth]{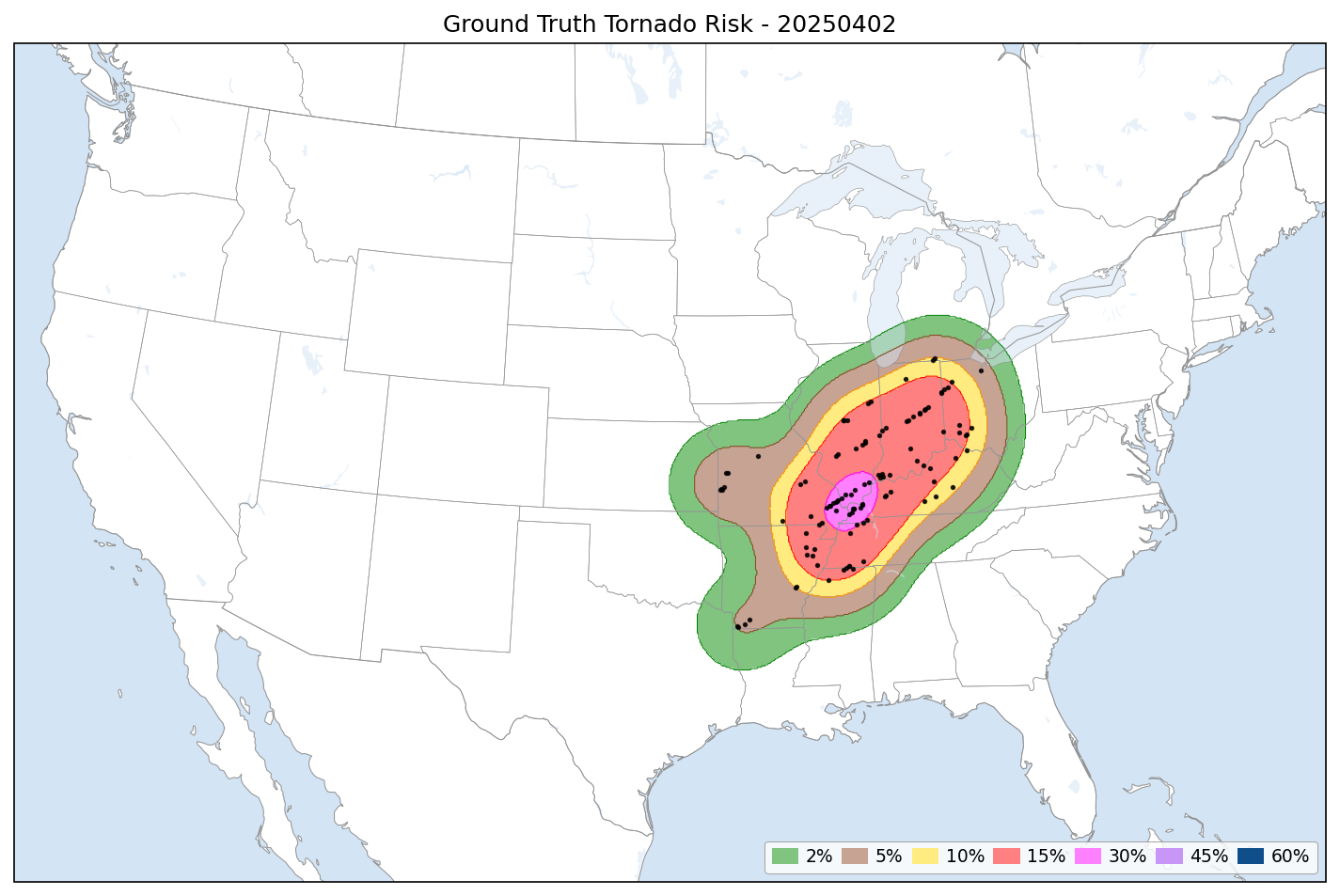}
    \end{minipage}
    \caption{Days with greater than 100 tornado reports.}
    \label{fig:100days}
\end{figure}

Converting discrete tornado reports into a continuous probability field requires spatial smoothing to capture the inherent uncertainty of tornado occurrences. To generate an objective verification target, we adapt and extend the Practically Perfect Forecast (PPF) methodology of \cite{hitchens_objective_2013}, developing a multi-step approach to construct high-resolution ground-truth risk fields. Our modified approach transforms discrete tornado observations into a continuous probability field representing a theoretically ideal probabilistic forecast, as displayed in Figure \ref{fig:100days}.

First, tornado reports from the SPC are aggregated for the relevant 24-hour forecast period (12:00 UTC to 12:00 UTC). A probability density field \( f(x, y) \) is calculated on an 80-km Lambert Conformal grid (NCEP Grid 211) using a normalized Gaussian kernel density estimator (KDE) with a smoothing parameter \( \sigma \approx 120 \text{ km} \):
\begin{equation}
f(x, y) = \sum_{n=1}^{N} \frac{1}{2\pi\sigma^2} \exp\left[ -\frac{1}{2} \left( \frac{d_n(x, y)}{\sigma} \right)^2 \right]
\end{equation}
where \(N\) is the total number of tornado reports, and \( d_n(x, y) \) is the Euclidean distance from grid point \((x, y)\) to the \(n\)-th report in the projected coordinate system. This density field \( f_{80km} \) is then bilinearly interpolated onto a finer grid with approximately 5-km spacing (\( f_{5km} \)), preserving the original projection. To align with the SPC's definition of tornado probability, \( f_{5km} \) is convolved with a uniform circular kernel of radius 40km. This integrates the probability density over the relevant neighborhood around each grid point. The result of the convolution is multiplied by the area of a 5-km grid cell (\( A_{\text{cell}} \)) to yield \( \lambda(x, y) \), the expected number of tornadoes in that neighborhood.
\begin{equation}
\lambda(x, y) = A_{\text{cell}} \cdot \text{Conv}(f_{5km}, \text{Disk}_{R=40km})(x, y)
\end{equation}
The ground truth probability is then calculated from this expected count via the Poisson relation.
\begin{equation}
P_{\text{truth}}(x, y) = 1 - e^{-\lambda(x, y)}
\end{equation}

The continuous \( P_{\text{truth}} \) field is categorized into discrete risk levels (`0\%', `2\%', `5\%', `10\%', `15\%', `30\%', `45\%', `60\%') based on standard SPC thresholds (e.g., \( 0.02 \le P_{\text{truth}}(i) < 0.05 \rightarrow \text{'2\%'} \)). These categorical raster areas are then converted into vector polygons; these polygons are reprojected from the Lambert Conformal grid CRS to standard geographic coordinates (WGS84) and saved as the daily ground truth file.

\subsection{TornadoBench Score}

We propose TornadoBench as the primary metric for AgentCaster. It is designed to evaluate the agent's ability to accurately delineate the location, extent, and intensity of tornado risk; it addresses the limitations of standard metrics by incorporating domain-specific weighting and geometric accuracy across multiple probability thresholds. For each day $d$ and risk category $C$ (from 0\% to 60\%), we calculate the IoU between the predicted and ground truth geometries where $\text{GT}_C$ and $\text{Pred}_C$ are the ground truth and predicted geometries for category $C$. For the 0\% category, we calculate the IoU of the complementary geometries. The daily TornadoBench score is then calculated as:

\begin{equation}
\text{TB}_d = 
\begin{cases} 
1 & \text{if } \text{MaxRisk}_{\text{GT},d} = 0\% \text{ and } \text{MaxRisk}_{\text{Pred},d} = 0\% \\ 
0 & \text{if } \text{MaxRisk}_{\text{GT},d} = 0\% \text{ and } \text{MaxRisk}_{\text{Pred},d} > 0\% \\ 
\frac{1}{|S_d|}\sum_{C \in S_d} \frac{\text{Area}(\text{GT}_C \cap \text{Pred}_C)}{\text{Area}(\text{GT}_C \cup \text{Pred}_C)} & \text{if } \text{MaxRisk}_{\text{GT},d} > 0\% 
\end{cases}
\end{equation}

where $|S_d|$ is the number of categories in set $S_d$. The overall TornadoBench score is a weighted average of daily scores, where the weight for each day depends on the maximum risk level in the ground truth:

\begin{equation}
\text{TornadoBench} = \frac{\sum_{d=1}^{D} (\text{TB}_d \cdot W_d)}{\sum_{d=1}^{D} W_d}
\end{equation}

where $W_d$ is the numerical value of the maximum risk level in the ground truth on day $d$ (e.g., `0\%' \(\rightarrow W_d=1\), `5\%' \(\rightarrow W_d=5\), `30\%' \(\rightarrow W_d=30\)).

\subsection{TornadoHallucination Metrics}

LLMs are known to hallucinate information \cite{zhang_sirens_2023, huang_survey_2025}, and in a forecasting context, we define this as predicting risk where none exists or predicting risk in an entirely non-overlapping location on a risk day. Evaluating hallucinations is particularly important in tornado prediction, where false alarms can lead to unnecessary costs and public complacency. We introduce two metrics to quantify these behaviors.

TornadoHallucinationSimple measures the frequency of simple false alarms: days where the agent predicted \textit{any} tornado risk (\( \text{MaxRisk}_{Pred, d} \ge 2\% \)) when the ground truth indicated \textit{no} risk (\( \text{MaxRisk}_{GT, d} = 0\% \)).

TornadoHallucinationHard penalizes hallucinations based on the \textit{magnitude} of incorrectly predicted risk. It considers two types of hallucinations: (1) any prediction of risk (\( \ge 2\% \)) on a quiet day (\( GT=0\% \)), and (2) predictions of risk (\( \ge 2\% \)) on a risk day (\( GT>0\% \)) that have \textit{zero spatial overlap} with the ground truth risk areas. Each such day is assigned a penalty equal to the numerical weight of the highest risk level predicted by the agent, as defined in TornadoBench. The final TornadoHallucinationHard score is computed as the average of these daily penalties over the benchmark period.

\subsection{Dataset Composition}
\label{dataset_composition}

The release benchmark dataset spans a continuous 40-day period from March 1, 2025, to April 9, 2025. This timeframe was selected to include a diverse range of meteorological scenarios across the CONUS, including quiet periods, marginal severe weather setups, and several significant tornado outbreak days. The overall distribution of maximum ground truth risk levels and associated tornado reports is summarized in Table~\ref{tab:dataset-composition-summary}. Detailed daily information, including the maximum ground truth risk, total tornado reports, and top affected states for each day in the benchmark period, is provided in Appendix~\ref{app:GT_details} (Table~\ref{tab:all_days_details}). While AgentCaster is designed for live daily forecasting, for benchmarking we select an evaluation window optimized for composition.

\begin{table}[h!]
  \caption{Distribution of maximum ground truth risk levels and associated tornado reports across the 40-day benchmark period (March 1--April 9, 2025). There were no 45\% or 60\% days.}
  \label{tab:dataset-composition-summary}
  \centering
  \begin{tabular}{lrr}
    \toprule
    Maximum Risk & Number of Days & Number of Reports \\
    \midrule
    0\%  & 22 & 5 \\
    2\%  & 6  & 21 \\
    5\%  & 4  & 44 \\
    10\% & 3  & 102 \\
    15\% & 2  & 45 \\
    30\% & 3  & 305 \\
    \midrule
    \textbf{Total} & \textbf{40} & \textbf{522} \\
    \bottomrule
  \end{tabular}
\end{table}

\section{Experiments and Evaluation}
\label{experiments}

We evaluated a suite of reasoning and non-reasoning multimodal LLMs with knowledge cutoff dates prior to March 1st. The human expert baseline is the first official SPC Day 1 Convective Outlook issued for the 12:00 UTC cycle, processed identically to agent predictions. All LLM agents were initialized with a detailed system prompt (see Appendix \ref{app:prompts_code} for full prompts) outlining their role as an AI meteorologist, the forecasting objective, data access tools, and the GeoJSON output format requirements.

\subsection{Main Results}

The primary forecasting accuracy, hallucination metrics, and maximum risk matching for the LLM configurations and the SPC baseline are presented in Table \ref{tab:main_performance_metrics_no_ablation_v2}. Agent interaction statistics and centroid distance errors are detailed in Table \ref{tab:interaction_and_distance_metrics_no_ablation} (centroid computation described in Appendix~\ref{app:centroid_calculations}). The SPC baseline achieves a TornadoBench score of 18.31\%, significantly outperforming all evaluated LLM agents. Among the LLM agents, performance varied, with the highest-scoring models achieving TornadoBench scores below 10\%. A notable challenge for several LLMs was the consistent generation of valid GeoJSON outputs. The models with the fewest valid predictions, gemini-2.5-flash-preview:thinking (16 days), also had the lowest TornadoBench scores.

Within the GPT-5 family, increasing reasoning correlates with a monotonic drop in TornadoBench (8.51\%, 7.23\%, 6.28\%, 3.54\% for gpt-5-minimal, gpt-5-low, gpt-5-medium, and gpt-5-high, respectively). This degradation occurs despite mixed shifts in hallucination severity. Furthermore, claude-3.7-sonnet (non-thinking) marginally outperforms its thinking variant on TornadoBench (6.79\% vs.\ 6.64\%).

\begin{table}[h!]
  \caption{Primary forecasting performance metrics. For TornadoHallucination metrics, lower is better. Max Risk Match shows the percentage of days the model's maximum predicted risk was Under/Match/Over the ground truth maximum risk.}
  \label{tab:main_performance_metrics_no_ablation_v2}
  \centering
  \resizebox{\textwidth}{!}{%
  \begin{tabular}{lccccc}
    \toprule
    Model & TornadoBench & TornadoHallucination & TornadoHallucination & \multicolumn{1}{c}{Max Risk Match (\%)} \\
    & (\%) & Simple & Hard & Under / Match / Over \\
    \midrule
    SPC (Human Expert) & 18.31 & 0.275 & 0.70 & 5.0 / 55.0 / 40.0 \\
    \midrule
    gpt-5-minimal & 8.51 & 0.385 & 2.56 & 12.8 / 20.5 / 66.7 \\
    gpt-5-low     & 7.23 & 0.444 & 1.92 & 11.1 / 27.8 / 61.1 \\
    claude-3.7-sonnet & 6.79 & 0.400 & 3.30 & 10.0 / 22.5 / 67.5 \\
    claude-3.7-sonnet:thinking & 6.64 & 0.359 & 3.10 & 17.9 / 23.1 / 59.0 \\
    gpt-5-medium  & 6.28 & 0.484 & 2.65 & 9.7 / 22.6 / 67.7 \\
    gpt-4.1 & 5.63 & 0.444 & 3.64 & 11.1 / 19.4 / 69.4 \\
    gemini-2.5-pro-preview-03-25 & 4.26 & 0.406 & 4.50 & 15.6 / 21.9 / 62.5 \\
    grok-4        & 3.85 & 0.538 & 8.85 & 2.6 / 7.7 / 89.7 \\
    gpt-5-high    & 3.54 & 0.500 & 2.30 & 16.7 / 0.0 / 83.3 \\
    o4-mini-high & 3.37 & 0.528 & 5.39 & 11.1 / 13.9 / 75.0 \\
    o3 & 3.27 & 0.550 & 5.50 & 10.0 / 7.5 / 82.5 \\
    gemini-2.5-flash-preview:thinking & 1.57 & 0.625 & 4.50 & 6.3 / 6.3 / 87.5 \\
    \bottomrule
  \end{tabular}%
  }
\end{table}

\begin{table}[h!]
  \caption{Agent interaction statistics and centroid distance errors.}
  \label{tab:interaction_and_distance_metrics_no_ablation}
  \centering
  \resizebox{\textwidth}{!}{%
  \begin{tabular}{lcccccc}
    \toprule
    Model & Prediction & Centroid Dist. & Avg. Assistant & Avg. Tool & Sounding Requests \\
    & Days & (Avg. / Max Risk) (km) & Turns & Calls & (Avg. / Max) \\
    \midrule
    SPC (Human Expert) & 40 & 182 / 236 & N/A & N/A & N/A \\
    \midrule
    gpt-5-minimal & 39 & 358 / 354 & 8.93 & 18.32 & 0.12 / 3 \\
    gpt-5-low     & 36 & 417 / 469 & 4.00 & 35.58 & 0.05 / 1 \\
    claude-3.7-sonnet & 40 & 405 / 441 & 21.80 & 21.80 & 4.83 / 8 \\
    claude-3.7-sonnet:thinking & 39 & 474 / 493 & 21.57 & 21.57 & 4.97 / 11 \\
    gpt-5-medium  & 31 & 398 / 447 & 4.45 & 41.27 & 0.05 / 1 \\
    gpt-4.1 & 36 & 361 / 377 & 11.32 & 23.07 & 4.47 / 13 \\
    gemini-2.5-pro-preview-03-25 & 32 & 494 / 561 & 5.55 & 18.38 & 2.23 / 5 \\
    grok-4        & 39 & 450 / 487 & 5.83 & 24.23 & 4.00 / 8 \\
    gpt-5-high    & 30 & 449 / 525 & 4.75 & 39.25 & 0.40 / 4 \\
    o4-mini-high & 36 & 583 / 623 & 6.58 & 6.55 & 0.12 / 1 \\
    o3 & 40 & 478 / 564 & 13.70 & 13.70 & 0.62 / 5 \\
    gemini-2.5-flash-preview:thinking & 16 & 601 / 595 & 7.05 & 32.38 & 2.70 / 50 \\
    \bottomrule
  \end{tabular}%
  }
\end{table}

LLM agents exhibit a strong tendency towards hallucinations. The TornadoHallucinationHard scores for LLMs were substantially higher than SPC's, with not only more frequent but also more severe hallucinations or complete misplacement of risk areas. The average centroid distance errors indicate significant challenges for LLMs in accurately placing the core of the predicted tornado threat, with most errors exceeding 400-500 km, compared to SPC's 182 km (overall) and 236 km (max risk). Agent interaction patterns varied across models. Except for one model, the number of sounding requests remained well below the daily quota of 50.

\subsection{High-Impact Tornado Outbreak Days}
\label{outbreak_analysis}

\begin{figure}[h!]
    \centering
    \includegraphics[width=0.99\linewidth]{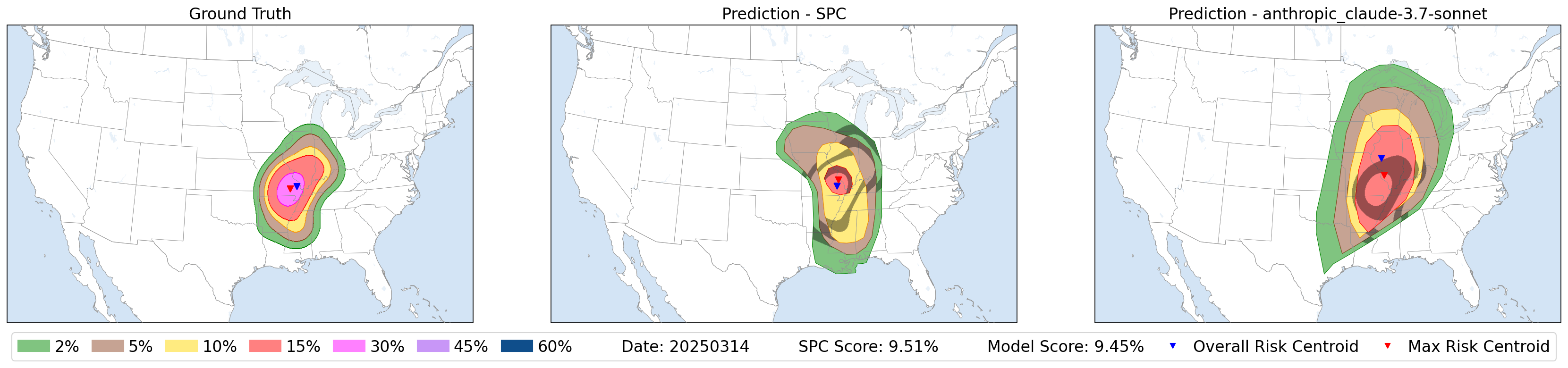}
    \caption{Evaluation of SPC and the top performing model on March 14, 2025. Overlapping solution regions are shaded.}
    \label{fig:100eval}
\end{figure}

Among the three 30\% risk days in our benchmark, we show March 14, 2025, the day whose SPC daily TornadoBench score is closest to the top model’s score. On this day, the top LLM agent achieved a daily TornadoBench score of 9.45\%, approaching SPC's 9.51\% (Figure \ref{fig:100eval}).

\section{Conclusion}

We introduced AgentCaster, a novel framework for evaluating multimodal LLM agents on the complex, real-world task of tornado forecasting. Through an interactive environment utilizing high-resolution meteorological data, AgentCaster assesses agentic reasoning in a high-impact domain. Our domain-specific metrics, TornadoBench and TornadoHallucination, applied over a 40-day period with significant severe weather, revealed substantial gaps between current LLM capabilities and human expert performance. Agents exhibited a strong tendency to hallucinate risk, overpredict its intensity, and struggled with precise geographic placement. By establishing a challenging benchmark in a high-stakes domain, we aim to drive progress toward more capable and reliable AI agents while simultaneously highlighting the current limitations of LLMs. The significant hallucination rates observed emphasize the need for continued research on model reliability before deployment in operational settings.


\printbibliography

@article{huang_survey_2025,
	title = {A Survey on Hallucination in Large Language Models: Principles, Taxonomy, Challenges, and Open Questions},
	volume = {43},
	issn = {1046-8188, 1558-2868},
	url = {http://arxiv.org/abs/2311.05232},
	doi = {10.1145/3703155},
	shorttitle = {A Survey on Hallucination in Large Language Models},
	pages = {1--55},
	number = {2},
	journaltitle = {{ACM} Transactions on Information Systems},
	shortjournal = {{ACM} Trans. Inf. Syst.},
	author = {Huang, Lei and Yu, Weijiang and Ma, Weitao and Zhong, Weihong and Feng, Zhangyin and Wang, Haotian and Chen, Qianglong and Peng, Weihua and Feng, Xiaocheng and Qin, Bing and Liu, Ting},
	urldate = {2025-05-05},
	date = {2025-03-31},
	eprinttype = {arxiv},
	eprint = {2311.05232 [cs]},
}

@misc{zhang_sirens_2023,
	title = {Siren's Song in the {AI} Ocean: A Survey on Hallucination in Large Language Models},
	url = {http://arxiv.org/abs/2309.01219},
	doi = {10.48550/arXiv.2309.01219},
	shorttitle = {Siren's Song in the {AI} Ocean},
	number = {{arXiv}:2309.01219},
	publisher = {{arXiv}},
	author = {Zhang, Yue and Li, Yafu and Cui, Leyang and Cai, Deng and Liu, Lemao and Fu, Tingchen and Huang, Xinting and Zhao, Enbo and Zhang, Yu and Chen, Yulong and Wang, Longyue and Luu, Anh Tuan and Bi, Wei and Shi, Freda and Shi, Shuming},
	urldate = {2025-05-05},
	date = {2023-09-24},
	eprinttype = {arxiv},
	eprint = {2309.01219 [cs]},
}

@article{powers_weather_2017,
	title = {The Weather Research and Forecasting Model: Overview, System Efforts, and Future Directions},
	url = {https://journals.ametsoc.org/view/journals/bams/98/8/bams-d-15-00308.1.xml},
	doi = {10.1175/BAMS-D-15-00308.1},
	shorttitle = {The Weather Research and Forecasting Model},
	author = {Powers, Jordan G. and Klemp, Joseph B. and Skamarock, William C. and Davis, Christopher A. and Dudhia, Jimy and Gill, David O. and Coen, Janice L. and Gochis, David J. and Ahmadov, Ravan and Peckham, Steven E. and Grell, Georg A. and Michalakes, John and Trahan, Samuel and Benjamin, Stanley G. and Alexander, Curtis R. and Dimego, Geoffrey J. and Wang, Wei and Schwartz, Craig S. and Romine, Glen S. and Liu, Zhiquan and Snyder, Chris and Chen, Fei and Barlage, Michael J. and Yu, Wei and Duda, Michael G.},
	urldate = {2025-05-04},
	date = {2017-08-01},
	langid = {english},
	note = {Section: Bulletin of the American Meteorological Society},
}

@article{blumberg_sharppy_2017,
	title = {{SHARPpy}: An Open-Source Sounding Analysis Toolkit for the Atmospheric Sciences},
	url = {https://journals.ametsoc.org/view/journals/bams/98/8/bams-d-15-00309.1.xml},
	doi = {10.1175/BAMS-D-15-00309.1},
	shorttitle = {{SHARPpy}},
	author = {Blumberg, William G. and Halbert, Kelton T. and Supinie, Timothy A. and Marsh, Patrick T. and Thompson, Richard L. and Hart, John A.},
	urldate = {2025-05-04},
	date = {2017-08-01},
	langid = {english},
	note = {Section: Bulletin of the American Meteorological Society},
}

@article{dowell_high-resolution_2022,
	title = {The High-Resolution Rapid Refresh ({HRRR}): An Hourly Updating Convection-Allowing Forecast Model. Part I: Motivation and System Description},
	url = {https://journals.ametsoc.org/view/journals/wefo/37/8/WAF-D-21-0151.1.xml},
	doi = {10.1175/WAF-D-21-0151.1},
	shorttitle = {The High-Resolution Rapid Refresh ({HRRR})},
	author = {Dowell, David C. and Alexander, Curtis R. and James, Eric P. and Weygandt, Stephen S. and Benjamin, Stanley G. and Manikin, Geoffrey S. and Blake, Benjamin T. and Brown, John M. and Olson, Joseph B. and Hu, Ming and Smirnova, Tatiana G. and Ladwig, Terra and Kenyon, Jaymes S. and Ahmadov, Ravan and Turner, David D. and Duda, Jeffrey D. and Alcott, Trevor I.},
	urldate = {2025-05-04},
	date = {2022-08-03},
	langid = {english},
	note = {Section: Weather and Forecasting},
}

@misc{yang_swe-agent_2024,
	title = {{SWE}-agent: Agent-Computer Interfaces Enable Automated Software Engineering},
	url = {http://arxiv.org/abs/2405.15793},
	doi = {10.48550/arXiv.2405.15793},
	shorttitle = {{SWE}-agent},
	number = {{arXiv}:2405.15793},
	publisher = {{arXiv}},
	author = {Yang, John and Jimenez, Carlos E. and Wettig, Alexander and Lieret, Kilian and Yao, Shunyu and Narasimhan, Karthik and Press, Ofir},
	urldate = {2025-04-29},
	date = {2024-11-11},
	eprinttype = {arxiv},
	eprint = {2405.15793 [cs]},
}

@misc{yang_dawn_2023,
	title = {The Dawn of {LMMs}: Preliminary Explorations with {GPT}-4V(ision)},
	url = {http://arxiv.org/abs/2309.17421},
	doi = {10.48550/arXiv.2309.17421},
	shorttitle = {The Dawn of {LMMs}},
	number = {{arXiv}:2309.17421},
	publisher = {{arXiv}},
	author = {Yang, Zhengyuan and Li, Linjie and Lin, Kevin and Wang, Jianfeng and Lin, Chung-Ching and Liu, Zicheng and Wang, Lijuan},
	urldate = {2025-04-29},
	date = {2023-10-11},
	eprinttype = {arxiv},
	eprint = {2309.17421 [cs]},
}

@misc{chen_are_2024,
	title = {Are We on the Right Way for Evaluating Large Vision-Language Models?},
	url = {http://arxiv.org/abs/2403.20330},
	doi = {10.48550/arXiv.2403.20330},
	number = {{arXiv}:2403.20330},
	publisher = {{arXiv}},
	author = {Chen, Lin and Li, Jinsong and Dong, Xiaoyi and Zhang, Pan and Zang, Yuhang and Chen, Zehui and Duan, Haodong and Wang, Jiaqi and Qiao, Yu and Lin, Dahua and Zhao, Feng},
	urldate = {2025-04-29},
	date = {2024-04-09},
	eprinttype = {arxiv},
	eprint = {2403.20330 [cs]},
}

@misc{zhu_minigpt-4_2023,
	title = {{MiniGPT}-4: Enhancing Vision-Language Understanding with Advanced Large Language Models},
	url = {http://arxiv.org/abs/2304.10592},
	doi = {10.48550/arXiv.2304.10592},
	shorttitle = {{MiniGPT}-4},
	number = {{arXiv}:2304.10592},
	publisher = {{arXiv}},
	author = {Zhu, Deyao and Chen, Jun and Shen, Xiaoqian and Li, Xiang and Elhoseiny, Mohamed},
	urldate = {2025-04-29},
	date = {2023-10-02},
	eprinttype = {arxiv},
	eprint = {2304.10592 [cs]},
}

@misc{radford_learning_2021,
	title = {Learning Transferable Visual Models From Natural Language Supervision},
	url = {http://arxiv.org/abs/2103.00020},
	doi = {10.48550/arXiv.2103.00020},
	number = {{arXiv}:2103.00020},
	publisher = {{arXiv}},
	author = {Radford, Alec and Kim, Jong Wook and Hallacy, Chris and Ramesh, Aditya and Goh, Gabriel and Agarwal, Sandhini and Sastry, Girish and Askell, Amanda and Mishkin, Pamela and Clark, Jack and Krueger, Gretchen and Sutskever, Ilya},
	urldate = {2025-04-29},
	date = {2021-02-26},
	eprinttype = {arxiv},
	eprint = {2103.00020 [cs]},
}

@article{zhang_vision-language_2024,
	title = {Vision-Language Models for Vision Tasks: A Survey},
	volume = {46},
	issn = {1939-3539},
	url = {https://ieeexplore.ieee.org/abstract/document/10445007},
	doi = {10.1109/TPAMI.2024.3369699},
	shorttitle = {Vision-Language Models for Vision Tasks},
	pages = {5625--5644},
	number = {8},
	journaltitle = {{IEEE} Transactions on Pattern Analysis and Machine Intelligence},
	author = {Zhang, Jingyi and Huang, Jiaxing and Jin, Sheng and Lu, Shijian},
	urldate = {2025-04-29},
	date = {2024-08},
}

@incollection{church_tornado_1993,
	location = {Washington, D. C.},
	title = {Tornado forecasting: A review},
	volume = {79},
	isbn = {978-0-87590-038-4},
	url = {http://doi.wiley.com/10.1029/GM079p0557},
	shorttitle = {Tornado forecasting},
	pages = {557--571},
	booktitle = {Geophysical Monograph Series},
	publisher = {American Geophysical Union},
	author = {Doswell, Charles A. and Weiss, Steven J. and Johns, Robert H.},
	editor = {Church, C. and Burgess, D. and Doswell, C. and Davies‐Jones, R.},
	urldate = {2025-04-29},
	date = {1993},
	langid = {english},
	doi = {10.1029/GM079p0557},
}

@article{lagerquist_deep_2020,
	title = {Deep Learning on Three-Dimensional Multiscale Data for Next-Hour Tornado Prediction},
	url = {https://journals.ametsoc.org/view/journals/mwre/148/7/mwrD190372.xml},
	doi = {10.1175/MWR-D-19-0372.1},
	author = {Lagerquist, Ryan and {McGovern}, Amy and Homeyer, Cameron R. and Ii, David John Gagne and Smith, Travis},
	urldate = {2025-04-29},
	date = {2020-06-24},
	langid = {english},
	note = {Section: Monthly Weather Review},
}

@misc{veillette_benchmark_2024,
	title = {A Benchmark Dataset for Tornado Detection and Prediction using Full-Resolution Polarimetric Weather Radar Data},
	url = {http://arxiv.org/abs/2401.16437},
	doi = {10.48550/arXiv.2401.16437},
	number = {{arXiv}:2401.16437},
	publisher = {{arXiv}},
	author = {Veillette, Mark S. and Kurdzo, James M. and Stepanian, Phillip M. and Cho, John Y. N. and Samsi, Siddharth and {McDonald}, Joseph},
	urldate = {2025-04-29},
	date = {2024-01-26},
	eprinttype = {arxiv},
	eprint = {2401.16437 [physics]},
}

@article{hill_forecasting_2020,
	title = {Forecasting Severe Weather with Random Forests},
	url = {https://journals.ametsoc.org/view/journals/mwre/148/5/mwr-d-19-0344.1.xml},
	doi = {10.1175/MWR-D-19-0344.1},
	author = {Hill, Aaron J. and Herman, Gregory R. and Schumacher, Russ S.},
	urldate = {2025-04-29},
	date = {2020-05-01},
	langid = {english},
	note = {Section: Monthly Weather Review},
}

@article{corfidi_birth_1999,
	title = {The Birth and Early Years of the Storm Prediction Center},
	issn = {1520-0434},
	url = {https://journals.ametsoc.org/view/journals/wefo/14/4/1520-0434_1999_014_0507_tbaeyo_2_0_co_2.xml},
	author = {Corfidi, Stephen F.},
	urldate = {2025-04-29},
	date = {1999-08-01},
	langid = {english},
	note = {Section: Weather and Forecasting},
}

@misc{wang_mint_2024,
	title = {{MINT}: Evaluating {LLMs} in Multi-turn Interaction with Tools and Language Feedback},
	url = {http://arxiv.org/abs/2309.10691},
	doi = {10.48550/arXiv.2309.10691},
	shorttitle = {{MINT}},
	number = {{arXiv}:2309.10691},
	publisher = {{arXiv}},
	author = {Wang, Xingyao and Wang, Zihan and Liu, Jiateng and Chen, Yangyi and Yuan, Lifan and Peng, Hao and Ji, Heng},
	urldate = {2025-04-29},
	date = {2024-03-12},
	eprinttype = {arxiv},
	eprint = {2309.10691 [cs]},
}

@misc{ma_agentboard_2024,
	title = {{AgentBoard}: An Analytical Evaluation Board of Multi-turn {LLM} Agents},
	url = {http://arxiv.org/abs/2401.13178},
	doi = {10.48550/arXiv.2401.13178},
	shorttitle = {{AgentBoard}},
	number = {{arXiv}:2401.13178},
	publisher = {{arXiv}},
	author = {Ma, Chang and Zhang, Junlei and Zhu, Zhihao and Yang, Cheng and Yang, Yujiu and Jin, Yaohui and Lan, Zhenzhong and Kong, Lingpeng and He, Junxian},
	urldate = {2025-04-29},
	date = {2024-12-23},
	eprinttype = {arxiv},
	eprint = {2401.13178 [cs]},
}

@misc{yi_survey_2024,
	title = {A Survey on Recent Advances in {LLM}-Based Multi-turn Dialogue Systems},
	url = {http://arxiv.org/abs/2402.18013},
	doi = {10.48550/arXiv.2402.18013},
	number = {{arXiv}:2402.18013},
	publisher = {{arXiv}},
	author = {Yi, Zihao and Ouyang, Jiarui and Liu, Yuwen and Liao, Tianhao and Xu, Zhe and Shen, Ying},
	urldate = {2025-04-29},
	date = {2024-02-28},
	eprinttype = {arxiv},
	eprint = {2402.18013 [cs]},
}

@article{wang_survey_2024,
	title = {A survey on large language model based autonomous agents},
	volume = {18},
	issn = {2095-2236},
	url = {https://doi.org/10.1007/s11704-024-40231-1},
	doi = {10.1007/s11704-024-40231-1},
	pages = {186345},
	number = {6},
	journaltitle = {Frontiers of Computer Science},
	shortjournal = {Front. Comput. Sci.},
	author = {Wang, Lei and Ma, Chen and Feng, Xueyang and Zhang, Zeyu and Yang, Hao and Zhang, Jingsen and Chen, Zhiyuan and Tang, Jiakai and Chen, Xu and Lin, Yankai and Zhao, Wayne Xin and Wei, Zhewei and Wen, Jirong},
	urldate = {2025-04-29},
	date = {2024-03-22},
	langid = {english},
}

@misc{naveed_comprehensive_2024,
	title = {A Comprehensive Overview of Large Language Models},
	url = {http://arxiv.org/abs/2307.06435},
	doi = {10.48550/arXiv.2307.06435},
	number = {{arXiv}:2307.06435},
	publisher = {{arXiv}},
	author = {Naveed, Humza and Khan, Asad Ullah and Qiu, Shi and Saqib, Muhammad and Anwar, Saeed and Usman, Muhammad and Akhtar, Naveed and Barnes, Nick and Mian, Ajmal},
	urldate = {2025-04-29},
	date = {2024-10-17},
	eprinttype = {arxiv},
	eprint = {2307.06435 [cs]},
}

@article{chang_survey_2024,
	title = {A Survey on Evaluation of Large Language Models},
	volume = {15},
	issn = {2157-6904},
	url = {https://dl.acm.org/doi/10.1145/3641289},
	doi = {10.1145/3641289},
	pages = {39:1--39:45},
	number = {3},
	journaltitle = {{ACM} Trans. Intell. Syst. Technol.},
	author = {Chang, Yupeng and Wang, Xu and Wang, Jindong and Wu, Yuan and Yang, Linyi and Zhu, Kaijie and Chen, Hao and Yi, Xiaoyuan and Wang, Cunxiang and Wang, Yidong and Ye, Wei and Zhang, Yue and Chang, Yi and Yu, Philip S. and Yang, Qiang and Xie, Xing},
	urldate = {2025-04-29},
	date = {2024-03-29},
}

@article{heinselman_warn--forecast_2023,
	title = {Warn-on-Forecast System: From Vision to Reality},
	url = {https://journals.ametsoc.org/view/journals/wefo/39/1/WAF-D-23-0147.1.xml},
	doi = {10.1175/WAF-D-23-0147.1},
	shorttitle = {Warn-on-Forecast System},
	author = {Heinselman, Pamela L. and Burke, Patrick C. and Wicker, Louis J. and Clark, Adam J. and Kain, John S. and Gao, Jidong and Yussouf, Nusrat and Jones, Thomas A. and Skinner, Patrick S. and Potvin, Corey K. and Wilson, Katie A. and Gallo, Burkely T. and Flora, Montgomery L. and Martin, Joshua and Creager, Gerry and Knopfmeier, Kent H. and Wang, Yunheng and Matilla, Brian C. and Dowell, David C. and Mansell, Edward R. and Roberts, Brett and Hoogewind, Kimberly A. and Stratman, Derek R. and Guerra, Jorge and Reinhart, Anthony E. and Kerr, Christopher A. and Miller, William},
	urldate = {2025-04-29},
	date = {2023-12-22},
	langid = {english},
	note = {Section: Weather and Forecasting},
}

@misc{bi_pangu-weather_2022,
	title = {Pangu-Weather: A 3D High-Resolution Model for Fast and Accurate Global Weather Forecast},
	url = {http://arxiv.org/abs/2211.02556},
	doi = {10.48550/arXiv.2211.02556},
	shorttitle = {Pangu-Weather},
	number = {{arXiv}:2211.02556},
	publisher = {{arXiv}},
	author = {Bi, Kaifeng and Xie, Lingxi and Zhang, Hengheng and Chen, Xin and Gu, Xiaotao and Tian, Qi},
	urldate = {2025-04-29},
	date = {2022-11-03},
	eprinttype = {arxiv},
	eprint = {2211.02556 [physics]},
}

@misc{pathak_fourcastnet_2022,
	title = {{FourCastNet}: A Global Data-driven High-resolution Weather Model using Adaptive Fourier Neural Operators},
	url = {http://arxiv.org/abs/2202.11214},
	doi = {10.48550/arXiv.2202.11214},
	shorttitle = {{FourCastNet}},
	number = {{arXiv}:2202.11214},
	publisher = {{arXiv}},
	author = {Pathak, Jaideep and Subramanian, Shashank and Harrington, Peter and Raja, Sanjeev and Chattopadhyay, Ashesh and Mardani, Morteza and Kurth, Thorsten and Hall, David and Li, Zongyi and Azizzadenesheli, Kamyar and Hassanzadeh, Pedram and Kashinath, Karthik and Anandkumar, Animashree},
	urldate = {2025-04-29},
	date = {2022-02-22},
	eprinttype = {arxiv},
	eprint = {2202.11214 [physics]},
}

@misc{lam_graphcast_2023,
	title = {{GraphCast}: Learning skillful medium-range global weather forecasting},
	url = {http://arxiv.org/abs/2212.12794},
	doi = {10.48550/arXiv.2212.12794},
	shorttitle = {{GraphCast}},
	number = {{arXiv}:2212.12794},
	publisher = {{arXiv}},
	author = {Lam, Remi and Sanchez-Gonzalez, Alvaro and Willson, Matthew and Wirnsberger, Peter and Fortunato, Meire and Alet, Ferran and Ravuri, Suman and Ewalds, Timo and Eaton-Rosen, Zach and Hu, Weihua and Merose, Alexander and Hoyer, Stephan and Holland, George and Vinyals, Oriol and Stott, Jacklynn and Pritzel, Alexander and Mohamed, Shakir and Battaglia, Peter},
	urldate = {2025-04-29},
	date = {2023-08-04},
	eprinttype = {arxiv},
	eprint = {2212.12794 [cs]},
}

@misc{xie_finben_2024,
	title = {{FinBen}: A Holistic Financial Benchmark for Large Language Models},
	url = {http://arxiv.org/abs/2402.12659},
	doi = {10.48550/arXiv.2402.12659},
	shorttitle = {{FinBen}},
	number = {{arXiv}:2402.12659},
	publisher = {{arXiv}},
	author = {Xie, Qianqian and Han, Weiguang and Chen, Zhengyu and Xiang, Ruoyu and Zhang, Xiao and He, Yueru and Xiao, Mengxi and Li, Dong and Dai, Yongfu and Feng, Duanyu and Xu, Yijing and Kang, Haoqiang and Kuang, Ziyan and Yuan, Chenhan and Yang, Kailai and Luo, Zheheng and Zhang, Tianlin and Liu, Zhiwei and Xiong, Guojun and Deng, Zhiyang and Jiang, Yuechen and Yao, Zhiyuan and Li, Haohang and Yu, Yangyang and Hu, Gang and Huang, Jiajia and Liu, Xiao-Yang and Lopez-Lira, Alejandro and Wang, Benyou and Lai, Yanzhao and Wang, Hao and Peng, Min and Ananiadou, Sophia and Huang, Jimin},
	urldate = {2025-04-29},
	date = {2024-06-19},
	eprinttype = {arxiv},
	eprint = {2402.12659 [cs]},
}

@misc{guha_legalbench_2023,
	title = {{LegalBench}: A Collaboratively Built Benchmark for Measuring Legal Reasoning in Large Language Models},
	url = {http://arxiv.org/abs/2308.11462},
	doi = {10.48550/arXiv.2308.11462},
	shorttitle = {{LegalBench}},
	number = {{arXiv}:2308.11462},
	publisher = {{arXiv}},
	author = {Guha, Neel and Nyarko, Julian and Ho, Daniel E. and Ré, Christopher and Chilton, Adam and Narayana, Aditya and Chohlas-Wood, Alex and Peters, Austin and Waldon, Brandon and Rockmore, Daniel N. and Zambrano, Diego and Talisman, Dmitry and Hoque, Enam and Surani, Faiz and Fagan, Frank and Sarfaty, Galit and Dickinson, Gregory M. and Porat, Haggai and Hegland, Jason and Wu, Jessica and Nudell, Joe and Niklaus, Joel and Nay, John and Choi, Jonathan H. and Tobia, Kevin and Hagan, Margaret and Ma, Megan and Livermore, Michael and Rasumov-Rahe, Nikon and Holzenberger, Nils and Kolt, Noam and Henderson, Peter and Rehaag, Sean and Goel, Sharad and Gao, Shang and Williams, Spencer and Gandhi, Sunny and Zur, Tom and Iyer, Varun and Li, Zehua},
	urldate = {2025-04-29},
	date = {2023-08-20},
	eprinttype = {arxiv},
	eprint = {2308.11462 [cs]},
}

@misc{yao_medqa-cs_2024,
	title = {{MedQA}-{CS}: Benchmarking Large Language Models Clinical Skills Using an {AI}-{SCE} Framework},
	url = {http://arxiv.org/abs/2410.01553},
	doi = {10.48550/arXiv.2410.01553},
	shorttitle = {{MedQA}-{CS}},
	number = {{arXiv}:2410.01553},
	publisher = {{arXiv}},
	author = {Yao, Zonghai and Zhang, Zihao and Tang, Chaolong and Bian, Xingyu and Zhao, Youxia and Yang, Zhichao and Wang, Junda and Zhou, Huixue and Jang, Won Seok and Ouyang, Feiyun and Yu, Hong},
	urldate = {2025-04-29},
	date = {2024-10-02},
	eprinttype = {arxiv},
	eprint = {2410.01553 [cs]},
}

@misc{singhal_large_2022,
	title = {Large Language Models Encode Clinical Knowledge},
	url = {http://arxiv.org/abs/2212.13138},
	doi = {10.48550/arXiv.2212.13138},
	number = {{arXiv}:2212.13138},
	publisher = {{arXiv}},
	author = {Singhal, Karan and Azizi, Shekoofeh and Tu, Tao and Mahdavi, S. Sara and Wei, Jason and Chung, Hyung Won and Scales, Nathan and Tanwani, Ajay and Cole-Lewis, Heather and Pfohl, Stephen and Payne, Perry and Seneviratne, Martin and Gamble, Paul and Kelly, Chris and Scharli, Nathaneal and Chowdhery, Aakanksha and Mansfield, Philip and Arcas, Blaise Aguera y and Webster, Dale and Corrado, Greg S. and Matias, Yossi and Chou, Katherine and Gottweis, Juraj and Tomasev, Nenad and Liu, Yun and Rajkomar, Alvin and Barral, Joelle and Semturs, Christopher and Karthikesalingam, Alan and Natarajan, Vivek},
	urldate = {2025-04-29},
	date = {2022-12-26},
	eprinttype = {arxiv},
	eprint = {2212.13138 [cs]},
}

@misc{jin_what_2020,
	title = {What Disease does this Patient Have? A Large-scale Open Domain Question Answering Dataset from Medical Exams},
	url = {http://arxiv.org/abs/2009.13081},
	doi = {10.48550/arXiv.2009.13081},
	shorttitle = {What Disease does this Patient Have?},
	number = {{arXiv}:2009.13081},
	publisher = {{arXiv}},
	author = {Jin, Di and Pan, Eileen and Oufattole, Nassim and Weng, Wei-Hung and Fang, Hanyi and Szolovits, Peter},
	urldate = {2025-04-29},
	date = {2020-09-28},
	eprinttype = {arxiv},
	eprint = {2009.13081 [cs]},
}

@misc{wu_st-think_2025,
	title = {{ST}-Think: How Multimodal Large Language Models Reason About 4D Worlds from Ego-Centric Videos},
	url = {http://arxiv.org/abs/2503.12542},
	doi = {10.48550/arXiv.2503.12542},
	shorttitle = {{ST}-Think},
	number = {{arXiv}:2503.12542},
	publisher = {{arXiv}},
	author = {Wu, Peiran and Liu, Yunze and Liu, Miao and Shen, Junxiao},
	urldate = {2025-04-29},
	date = {2025-04-23},
	eprinttype = {arxiv},
	eprint = {2503.12542 [cs]},
	note = {version: 2},
}

@misc{chen_rextime_2024,
	title = {{ReXTime}: A Benchmark Suite for Reasoning-Across-Time in Videos},
	url = {http://arxiv.org/abs/2406.19392},
	doi = {10.48550/arXiv.2406.19392},
	shorttitle = {{ReXTime}},
	number = {{arXiv}:2406.19392},
	publisher = {{arXiv}},
	author = {Chen, Jr-Jen and Liao, Yu-Chien and Lin, Hsi-Che and Yu, Yu-Chu and Chen, Yen-Chun and Wang, Yu-Chiang Frank},
	urldate = {2025-04-29},
	date = {2024-07-02},
	eprinttype = {arxiv},
	eprint = {2406.19392 [cs]},
}

@misc{su_living_2024,
	title = {Living in the Moment: Can Large Language Models Grasp Co-Temporal Reasoning?},
	url = {http://arxiv.org/abs/2406.09072},
	doi = {10.48550/arXiv.2406.09072},
	shorttitle = {Living in the Moment},
	number = {{arXiv}:2406.09072},
	publisher = {{arXiv}},
	author = {Su, Zhaochen and Li, Juntao and Zhang, Jun and Zhu, Tong and Qu, Xiaoye and Zhou, Pan and Bowen, Yan and Cheng, Yu and zhang, Min},
	urldate = {2025-04-29},
	date = {2024-06-13},
	eprinttype = {arxiv},
	eprint = {2406.09072 [cs]},
}

@misc{li_topviewrs_2024,
	title = {{TopViewRS}: Vision-Language Models as Top-View Spatial Reasoners},
	url = {http://arxiv.org/abs/2406.02537},
	doi = {10.48550/arXiv.2406.02537},
	shorttitle = {{TopViewRS}},
	number = {{arXiv}:2406.02537},
	publisher = {{arXiv}},
	author = {Li, Chengzu and Zhang, Caiqi and Zhou, Han and Collier, Nigel and Korhonen, Anna and Vulić, Ivan},
	urldate = {2025-04-29},
	date = {2024-06-04},
	eprinttype = {arxiv},
	eprint = {2406.02537 [cs]},
}

@misc{wang_is_2024,
	title = {Is A Picture Worth A Thousand Words? Delving Into Spatial Reasoning for Vision Language Models},
	url = {http://arxiv.org/abs/2406.14852},
	doi = {10.48550/arXiv.2406.14852},
	shorttitle = {Is A Picture Worth A Thousand Words?},
	number = {{arXiv}:2406.14852},
	publisher = {{arXiv}},
	author = {Wang, Jiayu and Ming, Yifei and Shi, Zhenmei and Vineet, Vibhav and Wang, Xin and Li, Yixuan and Joshi, Neel},
	urldate = {2025-04-29},
	date = {2024-11-04},
	eprinttype = {arxiv},
	eprint = {2406.14852 [cs]},
}

@misc{li_survey_2025,
	title = {A Survey of State of the Art Large Vision Language Models: Alignment, Benchmark, Evaluations and Challenges},
	url = {http://arxiv.org/abs/2501.02189},
	doi = {10.48550/arXiv.2501.02189},
	shorttitle = {A Survey of State of the Art Large Vision Language Models},
	number = {{arXiv}:2501.02189},
	publisher = {{arXiv}},
	author = {Li, Zongxia and Wu, Xiyang and Du, Hongyang and Liu, Fuxiao and Nghiem, Huy and Shi, Guangyao},
	urldate = {2025-04-29},
	date = {2025-04-06},
	eprinttype = {arxiv},
	eprint = {2501.02189 [cs]},
}

@misc{ruan_identifying_2024,
	title = {Identifying the Risks of {LM} Agents with an {LM}-Emulated Sandbox},
	url = {http://arxiv.org/abs/2309.15817},
	doi = {10.48550/arXiv.2309.15817},
	number = {{arXiv}:2309.15817},
	publisher = {{arXiv}},
	author = {Ruan, Yangjun and Dong, Honghua and Wang, Andrew and Pitis, Silviu and Zhou, Yongchao and Ba, Jimmy and Dubois, Yann and Maddison, Chris J. and Hashimoto, Tatsunori},
	urldate = {2025-04-29},
	date = {2024-05-17},
	eprinttype = {arxiv},
	eprint = {2309.15817 [cs]},
}

@misc{qin_toolllm_2023,
	title = {{ToolLLM}: Facilitating Large Language Models to Master 16000+ Real-world {APIs}},
	url = {http://arxiv.org/abs/2307.16789},
	doi = {10.48550/arXiv.2307.16789},
	shorttitle = {{ToolLLM}},
	number = {{arXiv}:2307.16789},
	publisher = {{arXiv}},
	author = {Qin, Yujia and Liang, Shihao and Ye, Yining and Zhu, Kunlun and Yan, Lan and Lu, Yaxi and Lin, Yankai and Cong, Xin and Tang, Xiangru and Qian, Bill and Zhao, Sihan and Hong, Lauren and Tian, Runchu and Xie, Ruobing and Zhou, Jie and Gerstein, Mark and Li, Dahai and Liu, Zhiyuan and Sun, Maosong},
	urldate = {2025-04-29},
	date = {2023-10-03},
	eprinttype = {arxiv},
	eprint = {2307.16789 [cs]},
}

@misc{zhou_webarena_2024,
	title = {{WebArena}: A Realistic Web Environment for Building Autonomous Agents},
	url = {http://arxiv.org/abs/2307.13854},
	doi = {10.48550/arXiv.2307.13854},
	shorttitle = {{WebArena}},
	number = {{arXiv}:2307.13854},
	publisher = {{arXiv}},
	author = {Zhou, Shuyan and Xu, Frank F. and Zhu, Hao and Zhou, Xuhui and Lo, Robert and Sridhar, Abishek and Cheng, Xianyi and Ou, Tianyue and Bisk, Yonatan and Fried, Daniel and Alon, Uri and Neubig, Graham},
	urldate = {2025-04-29},
	date = {2024-04-16},
	eprinttype = {arxiv},
	eprint = {2307.13854 [cs]},
}

@misc{liu_agentbench_2023,
	title = {{AgentBench}: Evaluating {LLMs} as Agents},
	url = {http://arxiv.org/abs/2308.03688},
	doi = {10.48550/arXiv.2308.03688},
	shorttitle = {{AgentBench}},
	number = {{arXiv}:2308.03688},
	publisher = {{arXiv}},
	author = {Liu, Xiao and Yu, Hao and Zhang, Hanchen and Xu, Yifan and Lei, Xuanyu and Lai, Hanyu and Gu, Yu and Ding, Hangliang and Men, Kaiwen and Yang, Kejuan and Zhang, Shudan and Deng, Xiang and Zeng, Aohan and Du, Zhengxiao and Zhang, Chenhui and Shen, Sheng and Zhang, Tianjun and Su, Yu and Sun, Huan and Huang, Minlie and Dong, Yuxiao and Tang, Jie},
	urldate = {2025-04-29},
	date = {2023-10-25},
	eprinttype = {arxiv},
	eprint = {2308.03688 [cs]},
}

@misc{phan_humanitys_2025,
	title = {Humanity's Last Exam},
	url = {http://arxiv.org/abs/2501.14249},
	doi = {10.48550/arXiv.2501.14249},
	number = {{arXiv}:2501.14249},
	publisher = {{arXiv}},
	author = {Phan, Long and Gatti, Alice and Han, Ziwen and Li, Nathaniel and Hu, Josephina and Zhang, Hugh and Zhang, Chen Bo Calvin and Shaaban, Mohamed and Ling, John and Shi, Sean and Choi, Michael and Agrawal, Anish and Chopra, Arnav and Khoja, Adam and Kim, Ryan and Ren, Richard and Hausenloy, Jason and Zhang, Oliver and Mazeika, Mantas and Dodonov, Dmitry and Nguyen, Tung and Lee, Jaeho and Anderson, Daron and Doroshenko, Mikhail and Stokes, Alun Cennyth and Mahmood, Mobeen and Pokutnyi, Oleksandr and Iskra, Oleg and Wang, Jessica P. and Levin, John-Clark and Kazakov, Mstyslav and Feng, Fiona and Feng, Steven Y. and Zhao, Haoran and Yu, Michael and Gangal, Varun and Zou, Chelsea and Wang, Zihan and Popov, Serguei and Gerbicz, Robert and Galgon, Geoff and Schmitt, Johannes and Yeadon, Will and Lee, Yongki and Sauers, Scott and Sanchez, Alvaro and Giska, Fabian and Roth, Marc and Riis, Søren and Utpala, Saiteja and Burns, Noah and Goshu, Gashaw M. and Naiya, Mohinder Maheshbhai and Agu, Chidozie and Giboney, Zachary and Cheatom, Antrell and Fournier-Facio, Francesco and Crowson, Sarah-Jane and Finke, Lennart and Cheng, Zerui and Zampese, Jennifer and Hoerr, Ryan G. and Nandor, Mark and Park, Hyunwoo and Gehrunger, Tim and Cai, Jiaqi and {McCarty}, Ben and Garretson, Alexis C. and Taylor, Edwin and Sileo, Damien and Ren, Qiuyu and Qazi, Usman and Li, Lianghui and Nam, Jungbae and Wydallis, John B. and Arkhipov, Pavel and Shi, Jack Wei Lun and Bacho, Aras and Willcocks, Chris G. and Cao, Hangrui and Motwani, Sumeet and Santos, Emily de Oliveira and Veith, Johannes and Vendrow, Edward and Cojoc, Doru and Zenitani, Kengo and Robinson, Joshua and Tang, Longke and Li, Yuqi and Vendrow, Joshua and Fraga, Natanael Wildner and Kuchkin, Vladyslav and Maksimov, Andrey Pupasov and Marion, Pierre and Efremov, Denis and Lynch, Jayson and Liang, Kaiqu and Mikov, Aleksandar and Gritsevskiy, Andrew and Guillod, Julien and Demir, Gözdenur and Martinez, Dakotah and Pageler, Ben and Zhou, Kevin and Soori, Saeed and Press, Ori and Tang, Henry and Rissone, Paolo and Green, Sean R. and Brüssel, Lina and Twayana, Moon and Dieuleveut, Aymeric and Imperial, Joseph Marvin and Prabhu, Ameya and Yang, Jinzhou and Crispino, Nick and Rao, Arun and Zvonkine, Dimitri and Loiseau, Gabriel and Kalinin, Mikhail and Lukas, Marco and Manolescu, Ciprian and Stambaugh, Nate and Mishra, Subrata and Hogg, Tad and Bosio, Carlo and Coppola, Brian P. and Salazar, Julian and Jin, Jaehyeok and Sayous, Rafael and Ivanov, Stefan and Schwaller, Philippe and Senthilkuma, Shaipranesh and Bran, Andres M. and Algaba, Andres and Houte, Kelsey Van den and Sypt, Lynn Van Der and Verbeken, Brecht and Noever, David and Kopylov, Alexei and Myklebust, Benjamin and Li, Bikun and Schut, Lisa and Zheltonozhskii, Evgenii and Yuan, Qiaochu and Lim, Derek and Stanley, Richard and Yang, Tong and Maar, John and Wykowski, Julian and Oller, Martí and Sahu, Anmol and Ardito, Cesare Giulio and Hu, Yuzheng and Kamdoum, Ariel Ghislain Kemogne and Jin, Alvin and Vilchis, Tobias Garcia and Zu, Yuexuan and Lackner, Martin and Koppel, James and Sun, Gongbo and Antonenko, Daniil S. and Chern, Steffi and Zhao, Bingchen and Arsene, Pierrot and Cavanagh, Joseph M. and Li, Daofeng and Shen, Jiawei and Crisostomi, Donato and Zhang, Wenjin and Dehghan, Ali and Ivanov, Sergey and Perrella, David and Kaparov, Nurdin and Zang, Allen and Sucholutsky, Ilia and Kharlamova, Arina and Orel, Daniil and Poritski, Vladislav and Ben-David, Shalev and Berger, Zachary and Whitfill, Parker and Foster, Michael and Munro, Daniel and Ho, Linh and Sivarajan, Shankar and Hava, Dan Bar and Kuchkin, Aleksey and Holmes, David and Rodriguez-Romero, Alexandra and Sommerhage, Frank and Zhang, Anji and Moat, Richard and Schneider, Keith and Kazibwe, Zakayo and Clarke, Don and Kim, Dae Hyun and Dias, Felipe Meneguitti and Fish, Sara and Elser, Veit and Kreiman, Tobias and Vilchis, Victor Efren Guadarrama and Klose, Immo and Anantheswaran, Ujjwala and Zweiger, Adam and Rawal, Kaivalya and Li, Jeffery and Nguyen, Jeremy and Daans, Nicolas and Heidinger, Haline and Radionov, Maksim and Rozhoň, Václav and Ginis, Vincent and Stump, Christian and Cohen, Niv and Poświata, Rafał and Tkadlec, Josef and Goldfarb, Alan and Wang, Chenguang and Padlewski, Piotr and Barzowski, Stanislaw and Montgomery, Kyle and Stendall, Ryan and Tucker-Foltz, Jamie and Stade, Jack and Rogers, T. Ryan and Goertzen, Tom and Grabb, Declan and Shukla, Abhishek and Givré, Alan and Ambay, John Arnold and Sen, Archan and Aziz, Muhammad Fayez and Inlow, Mark H. and He, Hao and Zhang, Ling and Kaddar, Younesse and Ängquist, Ivar and Chen, Yanxu and Wang, Harrison K. and Ramakrishnan, Kalyan and Thornley, Elliott and Terpin, Antonio and Schoelkopf, Hailey and Zheng, Eric and Carmi, Avishy and Brown, Ethan D. L. and Zhu, Kelin and Bartolo, Max and Wheeler, Richard and Stehberger, Martin and Bradshaw, Peter and Heimonen, J. P. and Sridhar, Kaustubh and Akov, Ido and Sandlin, Jennifer and Makarychev, Yury and Tam, Joanna and Hoang, Hieu and Cunningham, David M. and Goryachev, Vladimir and Patramanis, Demosthenes and Krause, Michael and Redenti, Andrew and Aldous, David and Lai, Jesyin and Coleman, Shannon and Xu, Jiangnan and Lee, Sangwon and Magoulas, Ilias and Zhao, Sandy and Tang, Ning and Cohen, Michael K. and Paradise, Orr and Kirchner, Jan Hendrik and Ovchynnikov, Maksym and Matos, Jason O. and Shenoy, Adithya and Wang, Michael and Nie, Yuzhou and Sztyber-Betley, Anna and Faraboschi, Paolo and Riblet, Robin and Crozier, Jonathan and Halasyamani, Shiv and Verma, Shreyas and Joshi, Prashant and Meril, Eli and Ma, Ziqiao and Andréoletti, Jérémy and Singhal, Raghav and Platnick, Jacob and Nevirkovets, Volodymyr and Basler, Luke and Ivanov, Alexander and Khoury, Seri and Gustafsson, Nils and Piccardo, Marco and Mostaghimi, Hamid and Chen, Qijia and Singh, Virendra and Khánh, Tran Quoc and Rosu, Paul and Szlyk, Hannah and Brown, Zachary and Narayan, Himanshu and Menezes, Aline and Roberts, Jonathan and Alley, William and Sun, Kunyang and Patel, Arkil and Lamparth, Max and Reuel, Anka and Xin, Linwei and Xu, Hanmeng and Loader, Jacob and Martin, Freddie and Wang, Zixuan and Achilleos, Andrea and Preu, Thomas and Korbak, Tomek and Bosio, Ida and Kazemi, Fereshteh and Chen, Ziye and Bálint, Biró and Lo, Eve J. Y. and Wang, Jiaqi and Nunes, Maria Inês S. and Milbauer, Jeremiah and Bari, M. Saiful and Wang, Zihao and Ansarinejad, Behzad and Sun, Yewen and Durand, Stephane and Elgnainy, Hossam and Douville, Guillaume and Tordera, Daniel and Balabanian, George and Wolff, Hew and Kvistad, Lynna and Milliron, Hsiaoyun and Sakor, Ahmad and Eron, Murat and O, Andrew Favre D. and Shah, Shailesh and Zhou, Xiaoxiang and Kamalov, Firuz and Abdoli, Sherwin and Santens, Tim and Barkan, Shaul and Tee, Allison and Zhang, Robin and Tomasiello, Alessandro and Luca, G. Bruno De and Looi, Shi-Zhuo and Le, Vinh-Kha and Kolt, Noam and Pan, Jiayi and Rodman, Emma and Drori, Jacob and Fossum, Carl J. and Muennighoff, Niklas and Jagota, Milind and Pradeep, Ronak and Fan, Honglu and Eicher, Jonathan and Chen, Michael and Thaman, Kushal and Merrill, William and Firsching, Moritz and Harris, Carter and Ciobâcă, Stefan and Gross, Jason and Pandey, Rohan and Gusev, Ilya and Jones, Adam and Agnihotri, Shashank and Zhelnov, Pavel and Mofayezi, Mohammadreza and Piperski, Alexander and Zhang, David K. and Dobarskyi, Kostiantyn and Leventov, Roman and Soroko, Ignat and Duersch, Joshua and Taamazyan, Vage and Ho, Andrew and Ma, Wenjie and Held, William and Xian, Ruicheng and Zebaze, Armel Randy and Mohamed, Mohanad and Leser, Julian Noah and Yuan, Michelle X. and Yacar, Laila and Lengler, Johannes and Olszewska, Katarzyna and Fratta, Claudio Di and Oliveira, Edson and Jackson, Joseph W. and Zou, Andy and Chidambaram, Muthu and Manik, Timothy and Haffenden, Hector and Stander, Dashiell and Dasouqi, Ali and Shen, Alexander and Golshani, Bita and Stap, David and Kretov, Egor and Uzhou, Mikalai and Zhidkovskaya, Alina Borisovna and Winter, Nick and Rodriguez, Miguel Orbegozo and Lauff, Robert and Wehr, Dustin and Tang, Colin and Hossain, Zaki and Phillips, Shaun and Samuele, Fortuna and Ekström, Fredrik and Hammon, Angela and Patel, Oam and Farhidi, Faraz and Medley, George and Mohammadzadeh, Forough and Peñaflor, Madellene and Kassahun, Haile and Friedrich, Alena and Perez, Rayner Hernandez and Pyda, Daniel and Sakal, Taom and Dhamane, Omkar and Mirabadi, Ali Khajegili and Hallman, Eric and Okutsu, Kenchi and Battaglia, Mike and Maghsoudimehrabani, Mohammad and Amit, Alon and Hulbert, Dave and Pereira, Roberto and Weber, Simon and Handoko and Peristyy, Anton and Malina, Stephen and Mehkary, Mustafa and Aly, Rami and Reidegeld, Frank and Dick, Anna-Katharina and Friday, Cary and Singh, Mukhwinder and Shapourian, Hassan and Kim, Wanyoung and Costa, Mariana and Gurdogan, Hubeyb and Kumar, Harsh and Ceconello, Chiara and Zhuang, Chao and Park, Haon and Carroll, Micah and Tawfeek, Andrew R. and Steinerberger, Stefan and Aggarwal, Daattavya and Kirchhof, Michael and Dai, Linjie and Kim, Evan and Ferret, Johan and Shah, Jainam and Wang, Yuzhou and Yan, Minghao and Burdzy, Krzysztof and Zhang, Lixin and Franca, Antonio and Pham, Diana T. and Loh, Kang Yong and Robinson, Joshua and Jackson, Abram and Giordano, Paolo and Petersen, Philipp and Cosma, Adrian and Colino, Jesus and White, Colin and Votava, Jacob and Vinnikov, Vladimir and Delaney, Ethan and Spelda, Petr and Stritecky, Vit and Shahid, Syed M. and Mourrat, Jean-Christophe and Vetoshkin, Lavr and Sponselee, Koen and Bacho, Renas and Yong, Zheng-Xin and Rosa, Florencia de la and Cho, Nathan and Li, Xiuyu and Malod, Guillaume and Weller, Orion and Albani, Guglielmo and Lang, Leon and Laurendeau, Julien and Kazakov, Dmitry and Adesanya, Fatimah and Portier, Julien and Hollom, Lawrence and Souza, Victor and Zhou, Yuchen Anna and Degorre, Julien and Yalın, Yiğit and Obikoya, Gbenga Daniel and Rai and Bigi, Filippo and Boscá, M. C. and Shumar, Oleg and Bacho, Kaniuar and Recchia, Gabriel and Popescu, Mara and Shulga, Nikita and Tanwie, Ngefor Mildred and Lux, Thomas C. H. and Rank, Ben and Ni, Colin and Brooks, Matthew and Yakimchyk, Alesia and Huanxu and Liu and Cavalleri, Stefano and Häggström, Olle and Verkama, Emil and Newbould, Joshua and Gundlach, Hans and Brito-Santana, Leonor and Amaro, Brian and Vajipey, Vivek and Grover, Rynaa and Wang, Ting and Kratish, Yosi and Li, Wen-Ding and Gopi, Sivakanth and Caciolai, Andrea and Witt, Christian Schroeder de and Hernández-Cámara, Pablo and Rodolà, Emanuele and Robins, Jules and Williamson, Dominic and Cheng, Vincent and Raynor, Brad and Qi, Hao and Segev, Ben and Fan, Jingxuan and Martinson, Sarah and Wang, Erik Y. and Hausknecht, Kaylie and Brenner, Michael P. and Mao, Mao and Demian, Christoph and Kassani, Peyman and Zhang, Xinyu and Avagian, David and Scipio, Eshawn Jessica and Ragoler, Alon and Tan, Justin and Sims, Blake and Plecnik, Rebeka and Kirtland, Aaron and Bodur, Omer Faruk and Shinde, D. P. and Labrador, Yan Carlos Leyva and Adoul, Zahra and Zekry, Mohamed and Karakoc, Ali and Santos, Tania C. B. and Shamseldeen, Samir and Karim, Loukmane and Liakhovitskaia, Anna and Resman, Nate and Farina, Nicholas and Gonzalez, Juan Carlos and Maayan, Gabe and Anderson, Earth and Pena, Rodrigo De Oliveira and Kelley, Elizabeth and Mariji, Hodjat and Pouriamanesh, Rasoul and Wu, Wentao and Finocchio, Ross and Alarab, Ismail and Cole, Joshua and Ferreira, Danyelle and Johnson, Bryan and Safdari, Mohammad and Dai, Liangti and Arthornthurasuk, Siriphan and {McAlister}, Isaac C. and Moyano, Alejandro José and Pronin, Alexey and Fan, Jing and Ramirez-Trinidad, Angel and Malysheva, Yana and Pottmaier, Daphiny and Taheri, Omid and Stepanic, Stanley and Perry, Samuel and Askew, Luke and Rodríguez, Raúl Adrián Huerta and Minissi, Ali M. R. and Lorena, Ricardo and Iyer, Krishnamurthy and Fasiludeen, Arshad Anil and Clark, Ronald and Ducey, Josh and Piza, Matheus and Somrak, Maja and Vergo, Eric and Qin, Juehang and Borbás, Benjámin and Chu, Eric and Lindsey, Jack and Jallon, Antoine and {McInnis}, I. M. J. and Chen, Evan and Semler, Avi and Gloor, Luk and Shah, Tej and Carauleanu, Marc and Lauer, Pascal and Huy, Tran Đuc and Shahrtash, Hossein and Duc, Emilien and Lewark, Lukas and Brown, Assaf and Albanie, Samuel and Weber, Brian and Vaz, Warren S. and Clavier, Pierre and Fan, Yiyang and Silva, Gabriel Poesia Reis e and Long and Lian and Abramovitch, Marcus and Jiang, Xi and Mendoza, Sandra and Islam, Murat and Gonzalez, Juan and Mavroudis, Vasilios and Xu, Justin and Kumar, Pawan and Goswami, Laxman Prasad and Bugas, Daniel and Heydari, Nasser and Jeanplong, Ferenc and Jansen, Thorben and Pinto, Antonella and Apronti, Archimedes and Galal, Abdallah and Ze-An, Ng and Singh, Ankit and Jiang, Tong and Xavier, Joan of Arc and Agarwal, Kanu Priya and Berkani, Mohammed and Zhang, Gang and Du, Zhehang and Junior, Benedito Alves de Oliveira and Malishev, Dmitry and Remy, Nicolas and Hartman, Taylor D. and Tarver, Tim and Mensah, Stephen and Loume, Gautier Abou and Morak, Wiktor and Habibi, Farzad and Hoback, Sarah and Cai, Will and Gimenez, Javier and Montecillo, Roselynn Grace and Łucki, Jakub and Campbell, Russell and Sharma, Asankhaya and Meer, Khalida and Gul, Shreen and Gonzalez, Daniel Espinosa and Alapont, Xavier and Hoover, Alex and Chhablani, Gunjan and Vargus, Freddie and Agarwal, Arunim and Jiang, Yibo and Patil, Deepakkumar and Outevsky, David and Scaria, Kevin Joseph and Maheshwari, Rajat and Dendane, Abdelkader and Shukla, Priti and Cartwright, Ashley and Bogdanov, Sergei and Mündler, Niels and Möller, Sören and Arnaboldi, Luca and Thaman, Kunvar and Siddiqi, Muhammad Rehan and Saxena, Prajvi and Gupta, Himanshu and Fruhauff, Tony and Sherman, Glen and Vincze, Mátyás and Usawasutsakorn, Siranut and Ler, Dylan and Radhakrishnan, Anil and Enyekwe, Innocent and Salauddin, Sk Md and Muzhen, Jiang and Maksapetyan, Aleksandr and Rossbach, Vivien and Harjadi, Chris and Bahaloohoreh, Mohsen and Sparrow, Claire and Sidhu, Jasdeep and Ali, Sam and Bian, Song and Lai, John and Singer, Eric and Uro, Justine Leon and Bateman, Greg and Sayed, Mohamed and Menshawy, Ahmed and Duclosel, Darling and Bezzi, Dario and Jain, Yashaswini and Aaron, Ashley and Tiryakioglu, Murat and Siddh, Sheeshram and Krenek, Keith and Shah, Imad Ali and Jin, Jun and Creighton, Scott and Peskoff, Denis and {EL}-Wasif, Zienab and V, Ragavendran P. and Richmond, Michael and {McGowan}, Joseph and Patwardhan, Tejal and Sun, Hao-Yu and Sun, Ting and Zubić, Nikola and Sala, Samuele and Ebert, Stephen and Kaddour, Jean and Schottdorf, Manuel and Wang, Dianzhuo and Petruzella, Gerol and Meiburg, Alex and Medved, Tilen and {ElSheikh}, Ali and Hebbar, S. Ashwin and Vaquero, Lorenzo and Yang, Xianjun and Poulos, Jason and Zouhar, Vilém and Bogdanik, Sergey and Zhang, Mingfang and Sanz-Ros, Jorge and Anugraha, David and Dai, Yinwei and Nhu, Anh N. and Wang, Xue and Demircali, Ali Anil and Jia, Zhibai and Zhou, Yuyin and Wu, Juncheng and He, Mike and Chandok, Nitin and Sinha, Aarush and Luo, Gaoxiang and Le, Long and Noyé, Mickaël and Perełkiewicz, Michał and Pantidis, Ioannis and Qi, Tianbo and Purohit, Soham Sachin and Parcalabescu, Letitia and Nguyen, Thai-Hoa and Winata, Genta Indra and Ponti, Edoardo M. and Li, Hanchen and Dhole, Kaustubh and Park, Jongee and Abbondanza, Dario and Wang, Yuanli and Nayak, Anupam and Caetano, Diogo M. and Wong, Antonio A. W. L. and Rio-Chanona, Maria del and Kondor, Dániel and Francois, Pieter and Chalstrey, Ed and Zsambok, Jakob and Hoyer, Dan and Reddish, Jenny and Hauser, Jakob and Rodrigo-Ginés, Francisco-Javier and Datta, Suchandra and Shepherd, Maxwell and Kamphuis, Thom and Zhang, Qizheng and Kim, Hyunjun and Sun, Ruiji and Yao, Jianzhu and Dernoncourt, Franck and Krishna, Satyapriya and Rismanchian, Sina and Pu, Bonan and Pinto, Francesco and Wang, Yingheng and Shridhar, Kumar and Overholt, Kalon J. and Briia, Glib and Nguyen, Hieu and David and Bartomeu, Soler and Pang, Tony {CY} and Wecker, Adam and Xiong, Yifan and Li, Fanfei and Huber, Lukas S. and Jaeger, Joshua and Maddalena, Romano De and Lù, Xing Han and Zhang, Yuhui and Beger, Claas and Kon, Patrick Tser Jern and Li, Sean and Sanker, Vivek and Yin, Ming and Liang, Yihao and Zhang, Xinlu and Agrawal, Ankit and Yifei, Li S. and Zhang, Zechen and Cai, Mu and Sonmez, Yasin and Cozianu, Costin and Li, Changhao and Slen, Alex and Yu, Shoubin and Park, Hyun Kyu and Sarti, Gabriele and Briański, Marcin and Stolfo, Alessandro and Nguyen, Truong An and Zhang, Mike and Perlitz, Yotam and Hernandez-Orallo, Jose and Li, Runjia and Shabani, Amin and Juefei-Xu, Felix and Dhingra, Shikhar and Zohar, Orr and Nguyen, My Chiffon and Pondaven, Alexander and Yilmaz, Abdurrahim and Zhao, Xuandong and Jin, Chuanyang and Jiang, Muyan and Todoran, Stefan and Han, Xinyao and Kreuer, Jules and Rabern, Brian and Plassart, Anna and Maggetti, Martino and Yap, Luther and Geirhos, Robert and Kean, Jonathon and Wang, Dingsu and Mollaei, Sina and Sun, Chenkai and Yin, Yifan and Wang, Shiqi and Li, Rui and Chang, Yaowen and Wei, Anjiang and Bizeul, Alice and Wang, Xiaohan and Arrais, Alexandre Oliveira and Mukherjee, Kushin and Chamorro-Padial, Jorge and Liu, Jiachen and Qu, Xingyu and Guan, Junyi and Bouyamourn, Adam and Wu, Shuyu and Plomecka, Martyna and Chen, Junda and Tang, Mengze and Deng, Jiaqi and Subramanian, Shreyas and Xi, Haocheng and Chen, Haoxuan and Zhang, Weizhi and Ren, Yinuo and Tu, Haoqin and Kim, Sejong and Chen, Yushun and Marjanović, Sara Vera and Ha, Junwoo and Luczyna, Grzegorz and Ma, Jeff J. and Shen, Zewen and Song, Dawn and Zhang, Cedegao E. and Wang, Zhun and Gendron, Gaël and Xiao, Yunze and Smucker, Leo and Weng, Erica and Lee, Kwok Hao and Ye, Zhe and Ermon, Stefano and Lopez-Miguel, Ignacio D. and Knights, Theo and Gitter, Anthony and Park, Namkyu and Wei, Boyi and Chen, Hongzheng and Pai, Kunal and Elkhanany, Ahmed and Lin, Han and Siedler, Philipp D. and Fang, Jichao and Mishra, Ritwik and Zsolnai-Fehér, Károly and Jiang, Xilin and Khan, Shadab and Yuan, Jun and Jain, Rishab Kumar and Lin, Xi and Peterson, Mike and Wang, Zhe and Malusare, Aditya and Tang, Maosen and Gupta, Isha and Fosin, Ivan and Kang, Timothy and Dworakowska, Barbara and Matsumoto, Kazuki and Zheng, Guangyao and Sewuster, Gerben and Villanueva, Jorge Pretel and Rannev, Ivan and Chernyavsky, Igor and Chen, Jiale and Banik, Deepayan and Racz, Ben and Dong, Wenchao and Wang, Jianxin and Bashmal, Laila and Gonçalves, Duarte V. and Hu, Wei and Bar, Kaushik and Bohdal, Ondrej and Patlan, Atharv Singh and Dhuliawala, Shehzaad and Geirhos, Caroline and Wist, Julien and Kansal, Yuval and Chen, Bingsen and Tire, Kutay and Yücel, Atak Talay and Christof, Brandon and Singla, Veerupaksh and Song, Zijian and Chen, Sanxing and Ge, Jiaxin and Ponkshe, Kaustubh and Park, Isaac and Shi, Tianneng and Ma, Martin Q. and Mak, Joshua and Lai, Sherwin and Moulin, Antoine and Cheng, Zhuo and Zhu, Zhanda and Zhang, Ziyi and Patil, Vaidehi and Jha, Ketan and Men, Qiutong and Wu, Jiaxuan and Zhang, Tianchi and Vieira, Bruno Hebling and Aji, Alham Fikri and Chung, Jae-Won and Mahfoud, Mohammed and Hoang, Ha Thi and Sperzel, Marc and Hao, Wei and Meding, Kristof and Xu, Sihan and Kostakos, Vassilis and Manini, Davide and Liu, Yueying and Toukmaji, Christopher and Paek, Jay and Yu, Eunmi and Demircali, Arif Engin and Sun, Zhiyi and Dewerpe, Ivan and Qin, Hongsen and Pflugfelder, Roman and Bailey, James and Morris, Johnathan and Heilala, Ville and Rosset, Sybille and Yu, Zishun and Chen, Peter E. and Yeo, Woongyeong and Jain, Eeshaan and Yang, Ryan and Chigurupati, Sreekar and Chernyavsky, Julia and Reddy, Sai Prajwal and Venugopalan, Subhashini and Batra, Hunar and Park, Core Francisco and Tran, Hieu and Maximiano, Guilherme and Zhang, Genghan and Liang, Yizhuo and Shiyu, Hu and Xu, Rongwu and Pan, Rui and Suresh, Siddharth and Liu, Ziqi and Gulati, Samaksh and Zhang, Songyang and Turchin, Peter and Bartlett, Christopher W. and Scotese, Christopher R. and Cao, Phuong M. and Nattanmai, Aakaash and {McKellips}, Gordon and Cheraku, Anish and Suhail, Asim and Luo, Ethan and Deng, Marvin and Luo, Jason and Zhang, Ashley and Jindel, Kavin and Paek, Jay and Halevy, Kasper and Baranov, Allen and Liu, Michael and Avadhanam, Advaith and Zhang, David and Cheng, Vincent and Ma, Brad and Fu, Evan and Do, Liam and Lass, Joshua and Yang, Hubert and Sunkari, Surya and Bharath, Vishruth and Ai, Violet and Leung, James and Agrawal, Rishit and Zhou, Alan and Chen, Kevin and Kalpathi, Tejas and Xu, Ziqi and Wang, Gavin and Xiao, Tyler and Maung, Erik and Lee, Sam and Yang, Ryan and Yue, Roy and Zhao, Ben and Yoon, Julia and Sun, Sunny and Singh, Aryan and Luo, Ethan and Peng, Clark and Osbey, Tyler and Wang, Taozhi and Echeazu, Daryl and Yang, Hubert and Wu, Timothy and Patel, Spandan and Kulkarni, Vidhi and Sundarapandiyan, Vijaykaarti and Zhang, Ashley and Le, Andrew and Nasim, Zafir and Yalam, Srikar and Kasamsetty, Ritesh and Samal, Soham and Yang, Hubert and Sun, David and Shah, Nihar and Saha, Abhijeet and Zhang, Alex and Nguyen, Leon and Nagumalli, Laasya and Wang, Kaixin and Zhou, Alan and Wu, Aidan and Luo, Jason and Telluri, Anwith and Yue, Summer and Wang, Alexandr and Hendrycks, Dan},
	urldate = {2025-04-29},
	date = {2025-04-19},
	eprinttype = {arxiv},
	eprint = {2501.14249 [cs]},
}

@misc{white_livebench_2025,
	title = {{LiveBench}: A Challenging, Contamination-Limited {LLM} Benchmark},
	url = {http://arxiv.org/abs/2406.19314},
	doi = {10.48550/arXiv.2406.19314},
	shorttitle = {{LiveBench}},
	number = {{arXiv}:2406.19314},
	publisher = {{arXiv}},
	author = {White, Colin and Dooley, Samuel and Roberts, Manley and Pal, Arka and Feuer, Ben and Jain, Siddhartha and Shwartz-Ziv, Ravid and Jain, Neel and Saifullah, Khalid and Dey, Sreemanti and Shubh-Agrawal and Sandha, Sandeep Singh and Naidu, Siddartha and Hegde, Chinmay and {LeCun}, Yann and Goldstein, Tom and Neiswanger, Willie and Goldblum, Micah},
	urldate = {2025-04-29},
	date = {2025-04-18},
	eprinttype = {arxiv},
	eprint = {2406.19314 [cs]},
}

@misc{chollet_arc_2025,
	title = {{ARC} Prize 2024: Technical Report},
	url = {http://arxiv.org/abs/2412.04604},
	doi = {10.48550/arXiv.2412.04604},
	shorttitle = {{ARC} Prize 2024},
	number = {{arXiv}:2412.04604},
	publisher = {{arXiv}},
	author = {Chollet, Francois and Knoop, Mike and Kamradt, Gregory and Landers, Bryan},
	urldate = {2025-04-29},
	date = {2025-01-08},
	eprinttype = {arxiv},
	eprint = {2412.04604 [cs]},
}

@misc{liang_holistic_2023,
	title = {Holistic Evaluation of Language Models},
	url = {http://arxiv.org/abs/2211.09110},
	doi = {10.48550/arXiv.2211.09110},
	number = {{arXiv}:2211.09110},
	publisher = {{arXiv}},
	author = {Liang, Percy and Bommasani, Rishi and Lee, Tony and Tsipras, Dimitris and Soylu, Dilara and Yasunaga, Michihiro and Zhang, Yian and Narayanan, Deepak and Wu, Yuhuai and Kumar, Ananya and Newman, Benjamin and Yuan, Binhang and Yan, Bobby and Zhang, Ce and Cosgrove, Christian and Manning, Christopher D. and Ré, Christopher and Acosta-Navas, Diana and Hudson, Drew A. and Zelikman, Eric and Durmus, Esin and Ladhak, Faisal and Rong, Frieda and Ren, Hongyu and Yao, Huaxiu and Wang, Jue and Santhanam, Keshav and Orr, Laurel and Zheng, Lucia and Yuksekgonul, Mert and Suzgun, Mirac and Kim, Nathan and Guha, Neel and Chatterji, Niladri and Khattab, Omar and Henderson, Peter and Huang, Qian and Chi, Ryan and Xie, Sang Michael and Santurkar, Shibani and Ganguli, Surya and Hashimoto, Tatsunori and Icard, Thomas and Zhang, Tianyi and Chaudhary, Vishrav and Wang, William and Li, Xuechen and Mai, Yifan and Zhang, Yuhui and Koreeda, Yuta},
	urldate = {2025-04-29},
	date = {2023-10-01},
	eprinttype = {arxiv},
	eprint = {2211.09110 [cs]},
}

@misc{srivastava_beyond_2023,
	title = {Beyond the Imitation Game: Quantifying and extrapolating the capabilities of language models},
	url = {http://arxiv.org/abs/2206.04615},
	doi = {10.48550/arXiv.2206.04615},
	shorttitle = {Beyond the Imitation Game},
	number = {{arXiv}:2206.04615},
	publisher = {{arXiv}},
	author = {Srivastava, Aarohi and Rastogi, Abhinav and Rao, Abhishek and Shoeb, Abu Awal Md and Abid, Abubakar and Fisch, Adam and Brown, Adam R. and Santoro, Adam and Gupta, Aditya and Garriga-Alonso, Adrià and Kluska, Agnieszka and Lewkowycz, Aitor and Agarwal, Akshat and Power, Alethea and Ray, Alex and Warstadt, Alex and Kocurek, Alexander W. and Safaya, Ali and Tazarv, Ali and Xiang, Alice and Parrish, Alicia and Nie, Allen and Hussain, Aman and Askell, Amanda and Dsouza, Amanda and Slone, Ambrose and Rahane, Ameet and Iyer, Anantharaman S. and Andreassen, Anders and Madotto, Andrea and Santilli, Andrea and Stuhlmüller, Andreas and Dai, Andrew and La, Andrew and Lampinen, Andrew and Zou, Andy and Jiang, Angela and Chen, Angelica and Vuong, Anh and Gupta, Animesh and Gottardi, Anna and Norelli, Antonio and Venkatesh, Anu and Gholamidavoodi, Arash and Tabassum, Arfa and Menezes, Arul and Kirubarajan, Arun and Mullokandov, Asher and Sabharwal, Ashish and Herrick, Austin and Efrat, Avia and Erdem, Aykut and Karakaş, Ayla and Roberts, B. Ryan and Loe, Bao Sheng and Zoph, Barret and Bojanowski, Bartłomiej and Özyurt, Batuhan and Hedayatnia, Behnam and Neyshabur, Behnam and Inden, Benjamin and Stein, Benno and Ekmekci, Berk and Lin, Bill Yuchen and Howald, Blake and Orinion, Bryan and Diao, Cameron and Dour, Cameron and Stinson, Catherine and Argueta, Cedrick and Ramírez, César Ferri and Singh, Chandan and Rathkopf, Charles and Meng, Chenlin and Baral, Chitta and Wu, Chiyu and Callison-Burch, Chris and Waites, Chris and Voigt, Christian and Manning, Christopher D. and Potts, Christopher and Ramirez, Cindy and Rivera, Clara E. and Siro, Clemencia and Raffel, Colin and Ashcraft, Courtney and Garbacea, Cristina and Sileo, Damien and Garrette, Dan and Hendrycks, Dan and Kilman, Dan and Roth, Dan and Freeman, Daniel and Khashabi, Daniel and Levy, Daniel and González, Daniel Moseguí and Perszyk, Danielle and Hernandez, Danny and Chen, Danqi and Ippolito, Daphne and Gilboa, Dar and Dohan, David and Drakard, David and Jurgens, David and Datta, Debajyoti and Ganguli, Deep and Emelin, Denis and Kleyko, Denis and Yuret, Deniz and Chen, Derek and Tam, Derek and Hupkes, Dieuwke and Misra, Diganta and Buzan, Dilyar and Mollo, Dimitri Coelho and Yang, Diyi and Lee, Dong-Ho and Schrader, Dylan and Shutova, Ekaterina and Cubuk, Ekin Dogus and Segal, Elad and Hagerman, Eleanor and Barnes, Elizabeth and Donoway, Elizabeth and Pavlick, Ellie and Rodola, Emanuele and Lam, Emma and Chu, Eric and Tang, Eric and Erdem, Erkut and Chang, Ernie and Chi, Ethan A. and Dyer, Ethan and Jerzak, Ethan and Kim, Ethan and Manyasi, Eunice Engefu and Zheltonozhskii, Evgenii and Xia, Fanyue and Siar, Fatemeh and Martínez-Plumed, Fernando and Happé, Francesca and Chollet, Francois and Rong, Frieda and Mishra, Gaurav and Winata, Genta Indra and Melo, Gerard de and Kruszewski, Germán and Parascandolo, Giambattista and Mariani, Giorgio and Wang, Gloria and Jaimovitch-López, Gonzalo and Betz, Gregor and Gur-Ari, Guy and Galijasevic, Hana and Kim, Hannah and Rashkin, Hannah and Hajishirzi, Hannaneh and Mehta, Harsh and Bogar, Hayden and Shevlin, Henry and Schütze, Hinrich and Yakura, Hiromu and Zhang, Hongming and Wong, Hugh Mee and Ng, Ian and Noble, Isaac and Jumelet, Jaap and Geissinger, Jack and Kernion, Jackson and Hilton, Jacob and Lee, Jaehoon and Fisac, Jaime Fernández and Simon, James B. and Koppel, James and Zheng, James and Zou, James and Kocoń, Jan and Thompson, Jana and Wingfield, Janelle and Kaplan, Jared and Radom, Jarema and Sohl-Dickstein, Jascha and Phang, Jason and Wei, Jason and Yosinski, Jason and Novikova, Jekaterina and Bosscher, Jelle and Marsh, Jennifer and Kim, Jeremy and Taal, Jeroen and Engel, Jesse and Alabi, Jesujoba and Xu, Jiacheng and Song, Jiaming and Tang, Jillian and Waweru, Joan and Burden, John and Miller, John and Balis, John U. and Batchelder, Jonathan and Berant, Jonathan and Frohberg, Jörg and Rozen, Jos and Hernandez-Orallo, Jose and Boudeman, Joseph and Guerr, Joseph and Jones, Joseph and Tenenbaum, Joshua B. and Rule, Joshua S. and Chua, Joyce and Kanclerz, Kamil and Livescu, Karen and Krauth, Karl and Gopalakrishnan, Karthik and Ignatyeva, Katerina and Markert, Katja and Dhole, Kaustubh D. and Gimpel, Kevin and Omondi, Kevin and Mathewson, Kory and Chiafullo, Kristen and Shkaruta, Ksenia and Shridhar, Kumar and {McDonell}, Kyle and Richardson, Kyle and Reynolds, Laria and Gao, Leo and Zhang, Li and Dugan, Liam and Qin, Lianhui and Contreras-Ochando, Lidia and Morency, Louis-Philippe and Moschella, Luca and Lam, Lucas and Noble, Lucy and Schmidt, Ludwig and He, Luheng and Colón, Luis Oliveros and Metz, Luke and Şenel, Lütfi Kerem and Bosma, Maarten and Sap, Maarten and Hoeve, Maartje ter and Farooqi, Maheen and Faruqui, Manaal and Mazeika, Mantas and Baturan, Marco and Marelli, Marco and Maru, Marco and Quintana, Maria Jose Ramírez and Tolkiehn, Marie and Giulianelli, Mario and Lewis, Martha and Potthast, Martin and Leavitt, Matthew L. and Hagen, Matthias and Schubert, Mátyás and Baitemirova, Medina Orduna and Arnaud, Melody and {McElrath}, Melvin and Yee, Michael A. and Cohen, Michael and Gu, Michael and Ivanitskiy, Michael and Starritt, Michael and Strube, Michael and Swędrowski, Michał and Bevilacqua, Michele and Yasunaga, Michihiro and Kale, Mihir and Cain, Mike and Xu, Mimee and Suzgun, Mirac and Walker, Mitch and Tiwari, Mo and Bansal, Mohit and Aminnaseri, Moin and Geva, Mor and Gheini, Mozhdeh and T, Mukund Varma and Peng, Nanyun and Chi, Nathan A. and Lee, Nayeon and Krakover, Neta Gur-Ari and Cameron, Nicholas and Roberts, Nicholas and Doiron, Nick and Martinez, Nicole and Nangia, Nikita and Deckers, Niklas and Muennighoff, Niklas and Keskar, Nitish Shirish and Iyer, Niveditha S. and Constant, Noah and Fiedel, Noah and Wen, Nuan and Zhang, Oliver and Agha, Omar and Elbaghdadi, Omar and Levy, Omer and Evans, Owain and Casares, Pablo Antonio Moreno and Doshi, Parth and Fung, Pascale and Liang, Paul Pu and Vicol, Paul and Alipoormolabashi, Pegah and Liao, Peiyuan and Liang, Percy and Chang, Peter and Eckersley, Peter and Htut, Phu Mon and Hwang, Pinyu and Miłkowski, Piotr and Patil, Piyush and Pezeshkpour, Pouya and Oli, Priti and Mei, Qiaozhu and Lyu, Qing and Chen, Qinlang and Banjade, Rabin and Rudolph, Rachel Etta and Gabriel, Raefer and Habacker, Rahel and Risco, Ramon and Millière, Raphaël and Garg, Rhythm and Barnes, Richard and Saurous, Rif A. and Arakawa, Riku and Raymaekers, Robbe and Frank, Robert and Sikand, Rohan and Novak, Roman and Sitelew, Roman and {LeBras}, Ronan and Liu, Rosanne and Jacobs, Rowan and Zhang, Rui and Salakhutdinov, Ruslan and Chi, Ryan and Lee, Ryan and Stovall, Ryan and Teehan, Ryan and Yang, Rylan and Singh, Sahib and Mohammad, Saif M. and Anand, Sajant and Dillavou, Sam and Shleifer, Sam and Wiseman, Sam and Gruetter, Samuel and Bowman, Samuel R. and Schoenholz, Samuel S. and Han, Sanghyun and Kwatra, Sanjeev and Rous, Sarah A. and Ghazarian, Sarik and Ghosh, Sayan and Casey, Sean and Bischoff, Sebastian and Gehrmann, Sebastian and Schuster, Sebastian and Sadeghi, Sepideh and Hamdan, Shadi and Zhou, Sharon and Srivastava, Shashank and Shi, Sherry and Singh, Shikhar and Asaadi, Shima and Gu, Shixiang Shane and Pachchigar, Shubh and Toshniwal, Shubham and Upadhyay, Shyam and Shyamolima and Debnath and Shakeri, Siamak and Thormeyer, Simon and Melzi, Simone and Reddy, Siva and Makini, Sneha Priscilla and Lee, Soo-Hwan and Torene, Spencer and Hatwar, Sriharsha and Dehaene, Stanislas and Divic, Stefan and Ermon, Stefano and Biderman, Stella and Lin, Stephanie and Prasad, Stephen and Piantadosi, Steven T. and Shieber, Stuart M. and Misherghi, Summer and Kiritchenko, Svetlana and Mishra, Swaroop and Linzen, Tal and Schuster, Tal and Li, Tao and Yu, Tao and Ali, Tariq and Hashimoto, Tatsu and Wu, Te-Lin and Desbordes, Théo and Rothschild, Theodore and Phan, Thomas and Wang, Tianle and Nkinyili, Tiberius and Schick, Timo and Kornev, Timofei and Tunduny, Titus and Gerstenberg, Tobias and Chang, Trenton and Neeraj, Trishala and Khot, Tushar and Shultz, Tyler and Shaham, Uri and Misra, Vedant and Demberg, Vera and Nyamai, Victoria and Raunak, Vikas and Ramasesh, Vinay and Prabhu, Vinay Uday and Padmakumar, Vishakh and Srikumar, Vivek and Fedus, William and Saunders, William and Zhang, William and Vossen, Wout and Ren, Xiang and Tong, Xiaoyu and Zhao, Xinran and Wu, Xinyi and Shen, Xudong and Yaghoobzadeh, Yadollah and Lakretz, Yair and Song, Yangqiu and Bahri, Yasaman and Choi, Yejin and Yang, Yichi and Hao, Yiding and Chen, Yifu and Belinkov, Yonatan and Hou, Yu and Hou, Yufang and Bai, Yuntao and Seid, Zachary and Zhao, Zhuoye and Wang, Zijian and Wang, Zijie J. and Wang, Zirui and Wu, Ziyi},
	urldate = {2025-04-29},
	date = {2023-06-12},
	eprinttype = {arxiv},
	eprint = {2206.04615 [cs]},
}

@misc{wang_mmlu-pro_2024,
	title = {{MMLU}-Pro: A More Robust and Challenging Multi-Task Language Understanding Benchmark},
	url = {http://arxiv.org/abs/2406.01574},
	doi = {10.48550/arXiv.2406.01574},
	shorttitle = {{MMLU}-Pro},
	number = {{arXiv}:2406.01574},
	publisher = {{arXiv}},
	author = {Wang, Yubo and Ma, Xueguang and Zhang, Ge and Ni, Yuansheng and Chandra, Abhranil and Guo, Shiguang and Ren, Weiming and Arulraj, Aaran and He, Xuan and Jiang, Ziyan and Li, Tianle and Ku, Max and Wang, Kai and Zhuang, Alex and Fan, Rongqi and Yue, Xiang and Chen, Wenhu},
	urldate = {2025-04-29},
	date = {2024-11-06},
	eprinttype = {arxiv},
	eprint = {2406.01574 [cs]},
}

@misc{hendrycks_measuring_2021,
	title = {Measuring Massive Multitask Language Understanding},
	url = {http://arxiv.org/abs/2009.03300},
	doi = {10.48550/arXiv.2009.03300},
	number = {{arXiv}:2009.03300},
	publisher = {{arXiv}},
	author = {Hendrycks, Dan and Burns, Collin and Basart, Steven and Zou, Andy and Mazeika, Mantas and Song, Dawn and Steinhardt, Jacob},
	urldate = {2025-04-29},
	date = {2021-01-12},
	eprinttype = {arxiv},
	eprint = {2009.03300 [cs]},
}

@misc{brown_language_2020,
	title = {Language Models are Few-Shot Learners},
	url = {http://arxiv.org/abs/2005.14165},
	doi = {10.48550/arXiv.2005.14165},
	number = {{arXiv}:2005.14165},
	publisher = {{arXiv}},
	author = {Brown, Tom B. and Mann, Benjamin and Ryder, Nick and Subbiah, Melanie and Kaplan, Jared and Dhariwal, Prafulla and Neelakantan, Arvind and Shyam, Pranav and Sastry, Girish and Askell, Amanda and Agarwal, Sandhini and Herbert-Voss, Ariel and Krueger, Gretchen and Henighan, Tom and Child, Rewon and Ramesh, Aditya and Ziegler, Daniel M. and Wu, Jeffrey and Winter, Clemens and Hesse, Christopher and Chen, Mark and Sigler, Eric and Litwin, Mateusz and Gray, Scott and Chess, Benjamin and Clark, Jack and Berner, Christopher and {McCandlish}, Sam and Radford, Alec and Sutskever, Ilya and Amodei, Dario},
	urldate = {2025-04-29},
	date = {2020-07-22},
	eprinttype = {arxiv},
	eprint = {2005.14165 [cs]},
}

@article{hitchens_objective_2013,
	title = {Objective Limits on Forecasting Skill of Rare Events},
	url = {https://journals.ametsoc.org/view/journals/wefo/28/2/waf-d-12-00113_1.xml},
	doi = {10.1175/WAF-D-12-00113.1},
	author = {Hitchens, Nathan M. and Brooks, Harold E. and Kay, Michael P.},
	urldate = {2025-04-27},
	date = {2013-04-01},
	langid = {english},
	note = {Section: Weather and Forecasting},
}

@article{lam_learning_2023,
	title = {Learning skillful medium-range global weather forecasting},
	volume = {382},
	url = {https://www.science.org/doi/10.1126/science.adi2336},
	doi = {10.1126/science.adi2336},
	pages = {1416--1421},
	number = {6677},
	journaltitle = {Science},
	author = {Lam, Remi and Sanchez-Gonzalez, Alvaro and Willson, Matthew and Wirnsberger, Peter and Fortunato, Meire and Alet, Ferran and Ravuri, Suman and Ewalds, Timo and Eaton-Rosen, Zach and Hu, Weihua and Merose, Alexander and Hoyer, Stephan and Holland, George and Vinyals, Oriol and Stott, Jacklynn and Pritzel, Alexander and Mohamed, Shakir and Battaglia, Peter},
	urldate = {2025-04-27},
	date = {2023-12-22},
	note = {Publisher: American Association for the Advancement of Science},
}

@online{noauthor_storm_nodate,
	title = {Storm Events Database {\textbar} National Centers for Environmental Information},
	url = {https://www.ncdc.noaa.gov/stormevents/},
	urldate = {2025-04-26},
}


\appendix

\newpage
\section{Limitations}
\label{app:limitations}

The evaluation period, while diverse, cannot capture the full range of meteorological conditions across multiple years. Current constraints of the benchmark include limiting sounding requests to 50 per day due to poor context handling and coherence loss with long contexts in existing models. With future models that demonstrate improved context window management and coherence, the framework could be easily extended to incorporate additional convection-allowing models or other data sources, as designed. The HRRRv4 remains the state-of-the-art convection-allowing model, providing both analyses (current conditions) and forecast maps \cite{powers_weather_2017}. Additionally, while we designed our interaction protocol to balance realism and reproducibility in the overall operational forecasting process, alternative approaches might better capture more specific aspects of the process. Future work could be to explore applications in related tasks, such as nowcasting or climate-scale forecasting.

\section{Prompts and Code}
\label{app:prompts_code}

The AgentCaster framework utilizes a structured prompting strategy to guide the LLM agent through the forecasting task. This includes an initial system prompt defining the agent's role, objectives, available tools, and evaluation criteria, followed by a first user prompt to initiate the interaction. The complete codebase for AgentCaster, including all code for data processing, agent interaction, and evaluation, as well as agent prediction GeoJSONs, is publicly available at \url{https://github.com/agentcaster/agentcaster}.

\subsection{System Prompt}

\begin{tcolorbox}[colback=blue!2!white,colframe=blue!60!black,title=AgentCaster System Prompt,fonttitle=\bfseries]
\textbf{You are AgentCaster, an expert autonomous AI meteorologist agent that issues Storm Prediction Center (SPC)-style forecasts in tornado prediction using 00z HRRR model data.}\\

\textbf{Objective:}\\
Your primary objective is to utilize HRRR forecast data to generate an SPC-style tornado risk forecast for the CONUS for the forecast day starting \texttt{\{date\_str\}} 12z to \texttt{\{next\_date\}} 12z (forecast hours 12-36 from the 00z run). This is the timeframe for which you will be making your SPC-style prediction.\\

\textbf{Background \& Evaluation:}\\
To evaluate your prediction, the ground truth is generated as follows: Observed tornado reports are used to calculate a normalized probability density field on an $\sim$80\,km grid (using a Gaussian kernel with $\sigma \approx 120\,\mathrm{km}$), which is then interpolated to a $\sim$5\,km grid. This density field is convolved with a 40\,km radius disk kernel to integrate the density over a neighborhood. The result is multiplied by the grid cell area to get an expected tornado count ($\lambda$). Finally, this expected count is converted to a probability using $P = 1 - e^{-\lambda}$. This probability field is categorized using standard SPC thresholds (2\%, 5\%, 10\%, etc.) and converted into vector polygon geometries. Your predicted risk areas (from the GeoJSON you provide) are directly compared against these ground truth geometries using vector-based geometric operations. Your final score is the average Intersection over Union (IoU) across all evaluated categories present in either your prediction or the ground truth, calculated based on the areas of the geometric intersection and union. This score ranges from 0\% (no agreement) to 100\% (perfect agreement). Accurate placement, spatial extent, and correct nesting of risk levels (2\%, 5\%, 10\%, etc.) are crucial for a high score. The tornado risk probabilities you predict (e.g., 5\%, 10\%) represent the likelihood of a tornado occurring within 25 miles (approx.\ 40\,km) of any point within that specific risk area during the forecast period (\texttt{\{date\_str\}} 12z to \texttt{\{next\_date\}} 12z).\\
\end{tcolorbox}

\begin{tcolorbox}[colback=purple!2!white,colframe=purple!60!black,title=Workflow Guidance,fonttitle=\bfseries]
- Start by calling \texttt{list\_available\_map\_types} to understand the data available for today.\\
- Then, use \texttt{request\_hrrr\_map} and \texttt{request\_sounding} (strategically, respecting the quota) to gather the information needed for your analysis.\\
- When confident, call \texttt{submit\_tornado\_prediction} with the properly formatted and nested GeoJSON output, ensuring all separate areas for each risk level are included.
\end{tcolorbox}

\begin{tcolorbox}[colback=purple!2!white,colframe=purple!60!black,title=Context \& Images,fonttitle=\bfseries]
- Map and sounding images are provided as PNGs embedded directly in the conversation (base64), and they consume context.\\
- [context limit provided to the model]
\end{tcolorbox}

\begin{tcolorbox}[colback=purple!2!white,colframe=purple!60!black,title=Autonomy,fonttitle=\bfseries]
- There is no human in the loop. Do not ask for permission or preferences.\\
- Decide and act yourself. If you need more evidence, request specific maps/soundings (respecting the quota and your context limit). Otherwise, proceed to call \texttt{submit\_tornado\_prediction} with a valid GeoJSON.\\
- Never ask questions like ``Which would you prefer?'' or ``Should I proceed?'' If you would ask, instead choose the action and perform it.
\end{tcolorbox}

\begin{tcolorbox}[colback=green!2!white,colframe=green!60!black,title=Tool: List Available Map Types,fonttitle=\bfseries]
\texttt{list\_available\_map\_types}:\\

Lists the available types of HRRR map plots based on the generated directories. Call this first to see what map types can be requested.
\end{tcolorbox}

\begin{tcolorbox}[colback=green!2!white,colframe=green!60!black,title=Tool: Request HRRR Map,fonttitle=\bfseries]
\texttt{request\_hrrr\_map}:\\

Requests a specific HRRR forecast map image (PNG). Provide the exact \texttt{map\_type\_directory} name from the list and the integer \texttt{forecast\_hour} (12--36).
\end{tcolorbox}

\begin{tcolorbox}[colback=white,colframe=gray!80!black,title=Required Properties,fonttitle=\bfseries]
- \textbf{map\_type\_directory} (\texttt{string}): The exact directory name representing the map type. \emph{Obtain this from \texttt{list\_available\_map\_types}.}\\
- \textbf{forecast\_hour} (\texttt{integer}): The forecast hour (e.g., 12, 18, 36) for the map.
\end{tcolorbox}

\begin{tcolorbox}[colback=green!2!white,colframe=green!60!black,title=Tool: Request Sounding,fonttitle=\bfseries]
\texttt{request\_sounding}:\\

Gets a sounding plot (PNG) for the nearest available station to a specified latitude and longitude for a specific integer \texttt{forecast\_hour} (12--36).\\
Limit of \texttt{\{max\_soundings\_per\_day\}} soundings per day.
\end{tcolorbox}

\begin{tcolorbox}[colback=white,colframe=gray!80!black,title=Required Properties,fonttitle=\bfseries]
- \textbf{latitude} (\texttt{number}): Target latitude in decimal degrees.\\
- \textbf{longitude} (\texttt{number}): Target longitude in decimal degrees.\\
- \textbf{forecast\_hour} (\texttt{integer}): Forecast hour (e.g., 12, 15, 24).
\end{tcolorbox}

\begin{tcolorbox}[colback=green!2!white,colframe=green!60!black,title=Tool: Submit Tornado Prediction,fonttitle=\bfseries]
\texttt{submit\_tornado\_prediction}:\\

Call this function \emph{only once} when you have finished analyzing all necessary maps and soundings and are ready to submit the final tornado risk prediction as GeoJSON.\\

\textbf{Format:}
The GeoJSON must be a valid \texttt{FeatureCollection} string representing the tornado risk forecast. Each feature must be a polygon or multipolygon with a \texttt{risk\_level} key in the \texttt{properties} field.\\

- Use \texttt{MultiPolygon} for disjoint risk areas.\\
- Ensure nesting: higher risk polygons must be spatially contained within all lower risk polygons.
\end{tcolorbox}

\begin{tcolorbox}[colback=white,colframe=gray!80!black,title=Required Properties,fonttitle=\bfseries]
- \textbf{prediction\_geojson} (\texttt{string}): Output GeoJSON \texttt{FeatureCollection} string as described above.
\end{tcolorbox}

\subsection{First User Prompt}

\begin{tcolorbox}[colback=orange!2!white,colframe=orange!60!black,title=First User Prompt,fonttitle=\bfseries]
Today's forecast date is \texttt{\{date\_str\}}.\\

You have \texttt{\{max\_soundings\_per\_day\}} sounding requests available for today.\\

Please start by calling \texttt{list\_available\_map\_types} to see the available map plots. Remember to call \texttt{submit\_tornado\_prediction} with your final GeoJSON prediction when you are confident with your analysis.
\end{tcolorbox}

\subsection{Context Limit Usage}

\begin{tcolorbox}[colback=cyan!2!white,colframe=cyan!60!black,title=Context Limit Usage (every turn),fonttitle=\bfseries]
Token usage: The current prompt is about [prompt tokens]. The conversation so far totals about [overall tokens]. [context limit provided to the model].
\end{tcolorbox}

\section{Centroid Calculation Methodology}
\label{app:centroid_calculations}

To complement the primary IoU-based TornadoBench score, we also calculate centroid-based metrics to quantify the geographic displacement of predicted tornado risk areas relative to the ground truth. These metrics capture both the central tendency of the overall risk and the core of the highest-threat regions. All centroid calculations are performed after reprojecting geometries to a common Lambert Conformal Conic projection (defined as \texttt{TARGET\_CRS}, based on NCEP Grid 211). Two primary types of centroids are computed for both the ground truth (GT) and the agent's prediction for each forecast day, as can be found in Figure \ref{fig:100eval}.

\paragraph{Overall risk centroid.} This centroid represents the geometric center of all areas where tornado risk $\geq 2\%$. For both GT and prediction, all individual disjoint risk polygons corresponding to risk levels of 2\% or greater are first combined into a single geometry using a \texttt{unary\_union} operation. This results in \texttt{geom\_gt\_nonzero} and \texttt{geom\_pred\_nonzero}, respectively. The centroid of this unified nonzero risk geometry is then calculated using the \texttt{.centroid} property of the resulting Shapely object, yielding $(x, y)$ coordinates in the \texttt{TARGET\_CRS}. This metric helps assess if the overall predicted envelope of tornado risk is geographically aligned with the observed risk envelope.

\paragraph{Maximum risk centroid.} This centroid represents the geometric center of the area(s) assigned the highest specific risk level present in the GT or prediction on a given day. For the GT, the maximum risk level observed on that day (e.g., ``30\%'') is identified (\texttt{current\_day\_max\_gt\_risk}). All disjoint polygons corresponding exclusively to this maximum risk level are combined using \texttt{unary\_union}. The centroid of this resulting geometry (\texttt{geom\_gt\_hr}) is then computed. For the prediction, the maximum risk level predicted by the agent (\texttt{max\_risk\_pred\_level}) is identified, and the centroid of the union of polygons for that specific highest predicted risk (\texttt{geom\_pred\_hr}) is calculated. This metric evaluates the agent's ability to pinpoint the core area of the most significant predicted or observed tornado threat.

\paragraph{Distance calculation.} Once the corresponding GT and predicted centroids (Overall Risk, Maximum Risk) are determined, the Euclidean distance between them is calculated. This distance is computed directly in the projected coordinate system (\texttt{TARGET\_CRS}), resulting in a value in meters. For reporting in summary tables and analyses, these distances are converted to kilometers (Table \ref{tab:interaction_and_distance_metrics_no_ablation}).

\section{Confidence Intervals}
\label{app:confidence}

Table \ref{tab:confidence_intervals} presents the $\pm 2 \sigma$ confidence intervals for key performance metrics. These intervals were calculated using a non-parametric bootstrap procedure with 1000 iterations for each model. The confidence intervals are derived using the percentile method from the distribution of bootstrap statistics; this method captures the variability in model performance due to the specific set of daily scores available for each model and is robust to non-normally distributed data, such that we allow for asymmetric intervals. Note that models evaluated on fewer prediction days may inherently exhibit wider confidence intervals due to a smaller sample size for bootstrapping.

\begin{table}[h!]
  \caption{Confidence intervals for performance metrics.}
  \label{tab:confidence_intervals}
  \centering
  \resizebox{\textwidth}{!}{%
  \begin{tabular}{lccc}
    \toprule
    Model & TornadoBench (\%) & TornadoHallucinationSimple & TornadoHallucinationHard \\
    \midrule
    SPC (Human Expert) & [10.23, 28.34] & [0.12, 0.42] & [0.30, 1.12] \\
    \midrule
    gpt-5-minimal & [4.80, 12.55] & [0.23, 0.54] & [1.58, 3.68] \\
    gpt-5-low     & [4.44, 12.32] & [0.28, 0.61] & [1.14, 2.81] \\
    claude-3.7-sonnet & [3.51, 10.78] & [0.25, 0.53] & [1.70, 4.97] \\
    claude-3.7-sonnet:thinking & [4.25, 10.61] & [0.21, 0.51] & [1.79, 4.56] \\
    gpt-5-medium  & [2.88, 11.21] & [0.29, 0.68] & [1.35, 4.27] \\
    gpt-4.1 & [3.58, 11.49] & [0.28, 0.61] & [2.22, 5.19] \\
    gemini-2.5-pro-preview-03-25 & [2.39, 6.94] & [0.25, 0.59] & [2.88, 6.09] \\
    grok-4        & [1.13, 7.67]  & [0.38, 0.69] & [6.28, 11.67] \\
    gpt-5-high    & [1.25, 7.77]  & [0.30, 0.67] & [1.39, 3.44] \\
    o4-mini-high & [1.70, 7.28] & [0.36, 0.69] & [3.94, 6.86] \\
    o3 & [1.14, 6.21] & [0.40, 0.71] & [3.84, 6.88] \\
    gemini-2.5-flash-preview:thinking & [0.34, 14.45] & [0.38, 0.88] & [2.44, 6.69] \\
    \bottomrule
  \end{tabular}%
  }
\end{table}

\section{Dataset Details}
\label{app:dataset_details}

The complete AgentCaster benchmark dataset, including all processed HRRR map types, soundings, and ground truths, is publicly available for research and reproducibility. The dataset is hosted on Hugging Face and can be accessed at \url{https://huggingface.co/datasets/agentcaster/agentcaster} ($\approx 244\,\mathrm{GB}$).

\subsection{NOAA License}

\begin{tcolorbox}[colback=gray!5!white, colframe=black, title=NOAA Data License]
NOAA data disseminated through NODD are open to the public and can be used as desired.\\

NOAA makes data openly available to ensure maximum use of our data, and to spur and encourage exploration and innovation throughout the industry. NOAA requests attribution for the use or dissemination of unaltered NOAA data. However, it is not permissible to state or imply endorsement by or affiliation with NOAA. If you modify NOAA data, you may not state or imply that it is original, unaltered NOAA data.

\medskip
\noindent Link: \url{https://registry.opendata.aws/noaa-hrrr-pds}.
\end{tcolorbox}

\subsection{List of Generated Forecast Maps}

\begin{longtable}{crl}
  \caption{List of the 141 map folders that were generated for the benchmark. While there are 145 available map types in the total data archive, some are organized within nested folders.} \\
  \toprule
  Var \# &  & Variable Name \\
  \midrule
  \endfirsthead

  \toprule
  Var \# &  & Variable Name \\
  \midrule
  \endhead

  \midrule
  \multicolumn{3}{r}{\textit{Continued on next page}} \\
  \bottomrule
  \endfoot

  \bottomrule
  \endlastfoot

    1  &  & 10\_metre\_U\_wind\_component\_at\_10\_heightAboveGround \\
    2  &  & 10\_metre\_V\_wind\_component\_at\_10\_heightAboveGround \\
    3  &  & 10\_metre\_wind\_speed\_at\_10\_heightAboveGround \\
    4  &  & 2\_metre\_dewpoint\_temperature\_at\_2\_heightAboveGround \\
    5  &  & 2\_metre\_relative\_humidity\_at\_2\_heightAboveGround \\
    6  &  & 2\_metre\_specific\_humidity\_at\_2\_heightAboveGround \\
    7  &  & 2\_metre\_temperature\_at\_2\_heightAboveGround \\
    8  &  & Aerosol\_optical\_depth\_at\_0\_atmosphereSingleLayer \\
    9  &  & Baseflow-groundwater\_runoff\_at\_0\_surface \\
    10 &  & Best\_(4-layer)\_lifted\_index\_at\_18000\_pressureFromGroundLayer\_Layer0Pa \\
    11 &  & Boundary\_layer\_height\_at\_0\_surface \\
    12 &  & Categorical\_freezing\_rain\_at\_0\_surface \\
    13 &  & Categorical\_ice\_pellets\_at\_0\_surface \\
    14 &  & Categorical\_rain\_at\_0\_surface \\
    15 &  & Categorical\_snow\_at\_0\_surface \\
    16 &  & Cloud\_Forcing\_Net\_Solar\_Flux\_at\_0\_surface \\
    17 &  & \makecell[l]{Convective\_available\_potential\_energy\_at\_0\_heightAboveGroundLayer\_ \\ Layer3000m} \\
    18 &  & Convective\_available\_potential\_energy\_at\_0\_surface \\
    19 &  & \makecell[l]{Convective\_available\_potential\_energy\_at\_18000 \\ pressureFromGroundLayer\_Layer0Pa} \\
    20 &  & \makecell[l]{Convective\_available\_potential\_energy\_at\_25500 \\ pressureFromGroundLayer\_Layer0Pa} \\
    21 &  & \makecell[l]{Convective\_available\_potential\_energy\_at\_9000 \\ pressureFromGroundLayer\_Layer0Pa} \\    
    22 &  & Convective\_inhibition\_at\_0\_surface \\
    23 &  & Convective\_inhibition\_at\_18000\_pressureFromGroundLayer\_Layer0Pa \\
    24 &  & Convective\_inhibition\_at\_25500\_pressureFromGroundLayer\_Layer0Pa \\
    25 &  & Convective\_inhibition\_at\_9000\_pressureFromGroundLayer\_Layer0Pa \\
    26 &  & Derived\_radar\_reflectivity\_at\_1000\_heightAboveGround \\
    27 &  & Derived\_radar\_reflectivity\_at\_263\_isothermal \\
    28 &  & Derived\_radar\_reflectivity\_at\_4000\_heightAboveGround \\
    29 &  & Dew\_point\_temperature\_at\_1000\_isobaricInhPa \\
    30 &  & Dew\_point\_temperature\_at\_500\_isobaricInhPa \\
    31 &  & Dew\_point\_temperature\_at\_700\_isobaricInhPa \\
    32 &  & Dew\_point\_temperature\_at\_850\_isobaricInhPa \\
    33 &  & Dew\_point\_temperature\_at\_925\_isobaricInhPa \\
    34 &  & Downward\_long-wave\_radiation\_flux\_at\_0\_surface \\
    35 &  & Downward\_short-wave\_radiation\_flux\_at\_0\_surface \\
    36 &  & Forecast\_surface\_roughness\_at\_0\_surface \\
    37 &  & Freezing\_Rain\_at\_0\_surface \\
    38 &  & Frictional\_velocity\_at\_0\_surface \\
    39 &  & Geometric\_vertical\_velocity\_at\_1\_sigmaLayer \\
    40 &  & Geometric\_vertical\_velocity\_at\_700\_isobaricInhPa \\
    41 &  & Geopotential\_height\_at\_0\_adiabaticCondensation \\
    42 &  & Geopotential\_height\_at\_0\_cloudBase \\
    43 &  & Geopotential\_height\_at\_0\_cloudCeiling \\
    44 &  & Geopotential\_height\_at\_0\_cloudTop \\
    45 &  & Geopotential\_height\_at\_0\_equilibrium \\
    46 &  & Geopotential\_height\_at\_0\_freeConvection \\
    47 &  & Geopotential\_height\_at\_0\_highestTroposphericFreezing \\
    48 &  & Geopotential\_height\_at\_0\_isothermZero \\
    49 &  & Geopotential\_height\_at\_1000\_isobaricInhPa \\
    50 &  & Geopotential\_height\_at\_253\_isothermal \\
    51 &  & Geopotential\_height\_at\_263\_isothermal \\
    52 &  & Geopotential\_height\_at\_500\_isobaricInhPa \\
    53 &  & Geopotential\_height\_at\_700\_isobaricInhPa \\
    54 &  & Geopotential\_height\_at\_850\_isobaricInhPa \\
    55 &  & Ground\_heat\_flux\_at\_0\_surface \\
    56 &  & Hail\_at\_0\_atmosphere \\
    57 &  & Hail\_at\_0\_sigma \\
    58 &  & Hail\_at\_0\_surface \\
    59 &  & High\_cloud\_cover\_at\_0\_highCloudLayer \\
    60 &  & Instantaneous\_surface\_sensible\_heat\_flux\_at\_0\_surface \\
    61 &  & Land-sea\_mask\_at\_0\_surface \\
    62 &  & Latent\_heat\_net\_flux\_at\_0\_surface \\
    63 &  & Layer\_Thickness\_261K-256K\_Layer \\
    64 &  & Leaf\_Area\_Index\_at\_0\_surface \\
    65 &  & Lightning\_at\_0\_atmosphere \\
    66 &  & Low\_cloud\_cover\_at\_0\_lowCloudLayer \\
    67 &  & Mass\_density\_at\_8\_heightAboveGround \\
    68 &  & Maximum\_Composite\_radar\_reflectivity\_at\_0\_atmosphere \\
    69 &  & Medium\_cloud\_cover\_at\_0\_middleCloudLayer \\
    70 &  & Moisture\_availability\_at\_0\_depthBelowLand \\
    71 &  & MSLP\_(MAPS\_System\_Reduction)\_at\_0\_meanSea \\
    72 &  & Orography\_at\_0\_surface \\
    73 &  & Percent\_frozen\_precipitation\_at\_0\_surface \\
    74 &  & Plant\_canopy\_surface\_water\_at\_0\_surface \\
    75 &  & Potential\_temperature\_at\_2\_heightAboveGround \\
    76 &  & Precipitable\_water\_at\_0\_atmosphereSingleLayer \\
    77 &  & Precipitation\_rate\_at\_0\_surface \\
    78 &  & Pressure\_at\_0\_cloudTop \\
    79 &  & Pressure\_at\_0\_highestTroposphericFreezing \\
    80 &  & Pressure\_at\_0\_isothermZero \\
    81 &  & Pressure\_at\_cloud\_base\_at\_0\_cloudBase \\
    82 &  & \makecell[l]{Pressure\_of\_level\_from\_which\_parcel\_was\_lifted\_at\_25500 \\ pressureFromGroundLayer\_Layer0Pa} \\
    83 &  & Relative\_humidity\_at\_0\_highestTroposphericFreezing \\
    84 &  & Relative\_humidity\_at\_0\_isothermZero \\
    85 &  & Sea\_ice\_area\_fraction\_at\_0\_surface \\
    86 &  & Simulated\_Brightness\_Temperature\_for\_GOES\_11,\_Channel\_3\_at\_0\_nominalTop \\
    87 &  & Simulated\_Brightness\_Temperature\_for\_GOES\_11,\_Channel\_4\_at\_0\_nominalTop \\
    88 &  & Simulated\_Brightness\_Temperature\_for\_GOES\_12,\_Channel\_3\_at\_0\_nominalTop \\
    89 &  & Simulated\_Brightness\_Temperature\_for\_GOES\_12,\_Channel\_4\_at\_0\_nominalTop \\
    90 &  & Snow\_cover\_at\_0\_surface \\
    91 &  & Snow\_depth\_at\_0\_surface \\
    92 &  & Storm\_relative\_helicity\_at\_1000\_heightAboveGroundLayer\_Layer0m \\
    93 &  & Storm\_relative\_helicity\_at\_3000\_heightAboveGroundLayer\_Layer0m \\
    94 &  & Storm\_surface\_runoff\_at\_0\_surface \\
    95 &  & Surface\_lifted\_index\_at\_500\_isobaricLayer\_Layer1000hPa \\
    96 &  & Surface\_pressure\_at\_0\_surface \\
    97 &  & Temperature\_at\_0\_surface \\
    98 &  & Temperature\_at\_1000\_isobaricInhPa \\
    99 &  & Temperature\_at\_500\_isobaricInhPa \\
    100 &  & Temperature\_at\_700\_isobaricInhPa \\
    101 &  & Temperature\_at\_850\_isobaricInhPa \\
    102 &  & Temperature\_at\_925\_isobaricInhPa \\
    103 &  & Total\_Cloud\_Cover\_at\_0\_atmosphere \\
    104 &  & Total\_Cloud\_Cover\_at\_0\_boundaryLayerCloudLayer \\
    105 &  & Total\_Precipitation\_at\_0\_surface \\
    106 &  & U\_component\_of\_wind\_at\_1000\_isobaricInhPa \\
    107 &  & U\_component\_of\_wind\_at\_250\_isobaricInhPa \\
    108 &  & U\_component\_of\_wind\_at\_300\_isobaricInhPa \\
    109 &  & U\_component\_of\_wind\_at\_500\_isobaricInhPa \\
    110 &  & U\_component\_of\_wind\_at\_700\_isobaricInhPa \\
    111 &  & U\_component\_of\_wind\_at\_80\_heightAboveGround \\
    112 &  & U\_component\_of\_wind\_at\_850\_isobaricInhPa \\
    113 &  & U\_component\_of\_wind\_at\_925\_isobaricInhPa \\
    114 &  & U-component\_storm\_motion\_at\_0\_heightAboveGroundLayer\_Layer6000m \\
    115 &  & Upward\_long-wave\_radiation\_flux\_at\_0\_nominalTop \\
    116 &  & Upward\_long-wave\_radiation\_flux\_at\_0\_surface \\
    117 &  & Upward\_short-wave\_radiation\_flux\_at\_0\_nominalTop \\
    118 &  & Upward\_short-wave\_radiation\_flux\_at\_0\_surface \\
    119 &  & V\_component\_of\_wind\_at\_1000\_isobaricInhPa \\
    120 &  & V\_component\_of\_wind\_at\_250\_isobaricInhPa \\
    121 &  & V\_component\_of\_wind\_at\_300\_isobaricInhPa \\
    122 &  & V\_component\_of\_wind\_at\_500\_isobaricInhPa \\
    123 &  & V\_component\_of\_wind\_at\_700\_isobaricInhPa \\
    124 &  & V\_component\_of\_wind\_at\_80\_heightAboveGround \\
    125 &  & V\_component\_of\_wind\_at\_850\_isobaricInhPa \\
    126 &  & V\_component\_of\_wind\_at\_925\_isobaricInhPa \\
    127 &  & V-component\_storm\_motion\_at\_0\_heightAboveGroundLayer\_Layer6000m \\
    128 &  & Vegetation\_at\_0\_surface \\
    129 &  & Vegetation\_Type\_at\_0\_surface \\
    130 &  & Vertical\_u-component\_shear\_at\_0\_heightAboveGroundLayer\_Layer1000m \\
    131 &  & Vertical\_u-component\_shear\_at\_0\_heightAboveGroundLayer\_Layer6000m \\
    132 &  & Vertical\_v-component\_shear\_at\_0\_heightAboveGroundLayer\_Layer1000m \\
    133 &  & Vertical\_v-component\_shear\_at\_0\_heightAboveGroundLayer\_Layer6000m \\
    134 &  & Vertically-integrated\_liquid\_at\_0\_atmosphere \\
    135 &  & Visibility\_at\_0\_surface \\
    136 &  & Visible\_Beam\_Downward\_Solar\_Flux\_at\_0\_surface \\
    137 &  & Visible\_Diffuse\_Downward\_Solar\_Flux\_at\_0\_surface \\
    138 &  & Vorticity\_(relative)\_at\_1000\_heightAboveGroundLayer\_Layer0m \\
    139 &  & Vorticity\_(relative)\_at\_2000\_heightAboveGroundLayer\_Layer0m \\
    140 &  & Water\_equivalent\_of\_accumulated\_snow\_depth\_(deprecated)\_at\_0\_surface \\
    141 &  & Wind\_speed\_(gust)\_at\_0\_surface \\
\end{longtable}

\section{Ground Truth Details}
\label{app:GT_details}

Table~\ref{tab:all_days_details} provides a day-by-day breakdown of the maximum ground truth tornado risk, the total number of observed tornado reports, and the top three states by report count for the entire 40-day benchmark period.

\begin{center}
\begin{longtable}{lcrl}
  \caption{Details for all 40 days in the benchmark period (March 1 -- April 9, 2025).} \label{tab:all_days_details} \\
  \toprule
  Date & Max Risk & Total Reports & Top 3 States (Report Count) \\
  \midrule
  \endfirsthead

  \toprule
  Date & Max Risk & Total Reports & Top 3 States (Report Count) \\
  \midrule
  \endhead

  \midrule
  \multicolumn{4}{r}{\textit{Continued on next page}} \\
  \bottomrule
  \endfoot

  \bottomrule
  \endlastfoot
    2025-03-01 & 0\% & 0 & N/A \\
    2025-03-02 & 0\% & 0 & N/A \\
    2025-03-03 & 2\% & 4 & TX (2), OK (2) \\
    2025-03-04 & 10\% & 26 & LA (10), TX (7), OK (6) \\
    2025-03-05 & 0\% & 2 & NC (1), VA (1) \\
    2025-03-06 & 0\% & 0 & N/A \\
    2025-03-07 & 0\% & 0 & N/A \\
    2025-03-08 & 0\% & 0 & N/A \\
    2025-03-09 & 0\% & 0 & N/A \\
    2025-03-10 & 0\% & 1 & FL (1) \\
    2025-03-11 & 0\% & 0 & N/A \\
    2025-03-12 & 0\% & 1 & CA (1) \\
    2025-03-13 & 0\% & 0 & N/A \\
    2025-03-14 & 30\% & 104 & MO (41), AR (21), IL (20) \\
    2025-03-15 & 30\% & 87 & MS (48), AL (26), LA (5) \\
    2025-03-16 & 5\% & 13 & PA (8), GA (2), NC (2) \\
    2025-03-17 & 0\% & 0 & N/A \\
    2025-03-18 & 0\% & 0 & N/A \\
    2025-03-19 & 15\% & 23 & IL (15), IN (7), KY (1) \\
    2025-03-20 & 0\% & 0 & N/A \\
    2025-03-21 & 0\% & 0 & N/A \\
    2025-03-22 & 0\% & 0 & N/A \\
    2025-03-23 & 2\% & 4 & MS (4) \\
    2025-03-24 & 0\% & 0 & N/A \\
    2025-03-25 & 0\% & 0 & N/A \\
    2025-03-26 & 0\% & 0 & N/A \\
    2025-03-27 & 2\% & 2 & TX (2) \\
    2025-03-28 & 2\% & 2 & TX (1), LA (1) \\
    2025-03-29 & 0\% & 1 & OK (1) \\
    2025-03-30 & 10\% & 51 & MI (14), IN (7), KY (7) \\
    2025-03-31 & 5\% & 9 & GA (5), AL (2), LA (1) \\
    2025-04-01 & 5\% & 9 & OK (5), KS (2), CA (1) \\
    2025-04-02 & 30\% & 114 & IN (22), MO (21), IL (21) \\
    2025-04-03 & 2\% & 5 & TN (2), KY (2), AL (1) \\
    2025-04-04 & 15\% & 22 & TX (15), AR (4), MO (2) \\
    2025-04-05 & 10\% & 25 & MS (16), TN (4), AL (4) \\
    2025-04-06 & 5\% & 13 & GA (7), AL (4), MS (2) \\
    2025-04-07 & 2\% & 4 & GA (3), FL (1) \\
    2025-04-08 & 0\% & 0 & N/A \\
    2025-04-09 & 0\% & 0 & N/A \\
\end{longtable}
\end{center}

\newpage
\section{TornadoBench Scores}

In the tables that follow, we first provide a concise mapping between each model’s internal identifier and its short-form abbreviation (Table \ref{tab:model-name-map}). Table \ref{tab:daily_tornadobench_scores} then presents the full set of daily TornadoBench scores for each model over the 40-day evaluation period; each row corresponds to one calendar date and dashed entries indicate days on which a model produced invalid GeoJSON output.

\begin{table}[h!]
  \caption{Model name mapping for Table \ref{tab:daily_tornadobench_scores}.}
  \label{tab:model-name-map}
  \centering
  \begin{tabular}{ll}
    \toprule
    Internal Name                                      & Abbreviation \\
    \midrule
    anthropic\_claude-3.7-sonnet                       & C3.7S  \\
    anthropic\_claude-3.7-sonnet:thinking              & C3.7ST \\
    google\_gemini-2.5-flash-preview:thinking          & G2.5FT \\
    google\_gemini-2.5-pro-preview-03-25               & G2.5P  \\
    openai\_gpt-4.1                                    & GPT4.1 \\
    openai\_o3                                         & O3     \\
    openai\_o4-mini-high                               & O4M    \\
    openai\_gpt-5-minimal              & GPT5   \\
    openai\_gpt-5-low                  & GPT5L  \\
    openai\_gpt-5-medium               & GPT5M  \\
    openai\_gpt-5-high                 & GPT5H  \\
    x-ai\_grok-4                        & G4     \\
    \bottomrule
  \end{tabular}
\end{table}

\begingroup
\tiny
\begin{longtable}[h!]{lrrrrrrrrrrrrr}
  \caption{Daily TornadoBench scores (rounded to the nearest percent). Dashed scores indicate invalid GeoJSONs.}
  \label{tab:daily_tornadobench_scores} \\
  \toprule
Date & SPC & C3.7S & C3.7ST & G2.5FT & G2.5P & G4 & GPT4.1 & GPT5 & GPT5H & GPT5L & GPT5M & O3 & O4M \\
  \midrule
  \endfirsthead

  \toprule
Date & SPC & C3.7S & C3.7ST & G2.5FT & G2.5P & G4 & GPT4.1 & GPT5 & GPT5H & GPT5L & GPT5M & O3 & O4M \\
  \midrule
  \endhead

  \midrule
  \multicolumn{10}{r}{\textit{Continued on next page}} \\
  \bottomrule
  \endfoot

  \bottomrule
  \endlastfoot
03-01 & 100\% & 0\% & 0\% & - & - & 0\% & 100\% & 0\% & 0\% & - & 0\% & 0\% & 0\% \\
03-02 & 0\% & 100\% & 100\% & - & 0\% & 0\% & 0\% & 0\% & 0\% & 0\% & - & 0\% & 0\% \\
03-03 & 3\% & 0\% & 0\% & 1\% & 0\% & 0\% & 1\% & 0\% & 4\% & 0\% & 0\% & 0\% & 2\% \\
03-04 & 10\% & 3\% & 5\% & - & 0\% & 5\% & 7\% & 7\% & 5\% & 7\% & 8\% & 4\% & 10\% \\
03-05 & 0\% & 0\% & 0\% & - & 0\% & 0\% & 0\% & 0\% & 0\% & 100\% & 0\% & 0\% & 0\% \\
03-06 & 100\% & 0\% & 100\% & 0\% & 0\% & 0\% & 0\% & 100\% & 0\% & 0\% & 0\% & 0\% & 100\% \\
03-07 & 100\% & 0\% & 0\% & 0\% & 0\% & 0\% & 0\% & 0\% & 0\% & 0\% & 0\% & 0\% & 0\% \\
03-08 & 0\% & 0\% & 100\% & 0\% & 0\% & 0\% & 0\% & 0\% & - & 0\% & - & 0\% & - \\
03-09 & 0\% & 0\% & 100\% & 0\% & - & 0\% & 0\% & 0\% & 0\% & 0\% & 0\% & 0\% & 0\% \\
03-10 & 0\% & 0\% & 0\% & - & 0\% & 0\% & 0\% & 0\% & 0\% & 100\% & 0\% & 0\% & 0\% \\
03-11 & 100\% & 100\% & 100\% & 100\% & 100\% & 0\% & 100\% & 100\% & - & 100\% & 0\% & 0\% & 0\% \\
03-12 & 0\% & 0\% & 100\% & 0\% & 0\% & - & 0\% & 0\% & - & 0\% & - & 0\% & 0\% \\
03-13 & 0\% & 100\% & 0\% & 0\% & 0\% & 0\% & - & - & 0\% & 0\% & 0\% & 0\% & 0\% \\
03-14 & 10\% & 9\% & 4\% & - & 3\% & 4\% & - & 6\% & 5\% & 7\% & 7\% & 3\% & 3\% \\
03-15 & 18\% & 0\% & 6\% & 6\% & 3\% & 6\% & 0\% & 8\% & 7\% & 5\% & 11\% & 7\% & 4\% \\
03-16 & 10\% & 1\% & 0\% & 0\% & 1\% & 0\% & - & 3\% & 2\% & 7\% & - & 0\% & 0\% \\
03-17 & 100\% & 100\% & 0\% & 0\% & - & 0\% & 100\% & 100\% & 0\% & 0\% & 100\% & 0\% & 0\% \\
03-18 & 100\% & 0\% & 0\% & - & - & 0\% & 100\% & 100\% & 0\% & 100\% & 0\% & 0\% & 0\% \\
03-19 & 23\% & 2\% & - & - & 8\% & 4\% & 0\% & 10\% & 3\% & - & 1\% & 2\% & 0\% \\
03-20 & 100\% & 100\% & 100\% & 0\% & 0\% & 0\% & 0\% & 0\% & 0\% & 0\% & 0\% & 0\% & 0\% \\
03-21 & 100\% & 0\% & 0\% & 0\% & 0\% & 0\% & 0\% & 0\% & 0\% & 0\% & 0\% & 0\% & 0\% \\
03-22 & 0\% & 0\% & 0\% & - & - & 0\% & 0\% & 0\% & - & 0\% & - & 0\% & 0\% \\
03-23 & 6\% & 2\% & 0\% & - & 0\% & 0\% & 0\% & 0\% & 0\% & 0\% & 0\% & 0\% & 0\% \\
03-24 & 0\% & 0\% & 0\% & - & 0\% & 0\% & 0\% & 0\% & 0\% & 0\% & - & 0\% & 0\% \\
03-25 & 100\% & 0\% & 0\% & - & - & 0\% & 0\% & 100\% & - & 0\% & 0\% & 0\% & 0\% \\
03-26 & 0\% & 100\% & 100\% & - & 0\% & 0\% & 0\% & 0\% & - & 0\% & 0\% & 0\% & 0\% \\
03-27 & 11\% & 0\% & 0\% & - & 0\% & 1\% & 0\% & 0\% & 0\% & 0\% & 0\% & 0\% & 0\% \\
03-28 & 0\% & 0\% & 0\% & - & - & 0\% & 0\% & 0\% & 0\% & 0\% & - & 0\% & 0\% \\
03-29 & 0\% & 0\% & 0\% & - & 0\% & 0\% & 0\% & 0\% & - & 0\% & 0\% & 0\% & 0\% \\
03-30 & 15\% & 3\% & 2\% & - & 7\% & 3\% & - & 3\% & - & 2\% & 8\% & 2\% & - \\
03-31 & 4\% & 4\% & 3\% & - & 6\% & 1\% & 0\% & 9\% & - & 10\% & 4\% & 1\% & 3\% \\
04-01 & 2\% & 0\% & 0\% & - & 0\% & 2\% & 3\% & 5\% & 3\% & - & 2\% & 1\% & - \\
04-02 & 23\% & 7\% & 1\% & - & 3\% & 8\% & 7\% & 6\% & 5\% & 6\% & 5\% & 4\% & - \\
04-03 & 2\% & 0\% & 0\% & 5\% & 0\% & 0\% & 3\% & 0\% & 0\% & 0\% & 0\% & 0\% & 3\% \\
04-04 & 13\% & 5\% & 6\% & - & 2\% & 2\% & 7\% & 4\% & 0\% & 4\% & 3\% & 4\% & 5\% \\
04-05 & 7\% & 6\% & 5\% & - & 3\% & 1\% & 12\% & 12\% & - & 18\% & 13\% & 6\% & 7\% \\
04-06 & 26\% & 6\% & 7\% & 3\% & 6\% & 3\% & 19\% & 4\% & 13\% & 5\% & - & 3\% & 1\% \\
04-07 & 4\% & 0\% & 0\% & - & - & 2\% & 3\% & 0\% & 7\% & - & - & 0\% & 0\% \\
04-08 & 100\% & 0\% & 0\% & - & 100\% & 0\% & 0\% & 100\% & 0\% & 100\% & 100\% & 0\% & 0\% \\
04-09 & 100\% & 0\% & 0\% & 0\% & 100\% & 0\% & 100\% & 0\% & 0\% & 0\% & 0\% & 0\% & 100\% \\
\end{longtable}
\endgroup

\section{LLM Reasoning Details}
\label{LLM_details}

\begin{table}[H]
  \caption{Reasoning capabilities of evaluated models. Parentheses indicate maximum length of reasoning (tokens) if relevant.}
  \label{tab:models-evaluated}
  \centering
  \begin{tabular}{lll}
    \toprule
    Provider  & Model Name                           & Reasoning \\
    \midrule
    Anthropic & claude-3.7-sonnet                    & No        \\
    Anthropic & claude-3.7-sonnet\texttt{:}thinking  & Yes (32k) \\
    Google    & gemini-2.5-flash-preview\texttt{:}thinking & Yes (25k) \\
    Google    & gemini-2.5-pro-preview-03-25         & Yes       \\
    OpenAI    & gpt-4.1                              & No        \\
    OpenAI    & o3                                   & Yes       \\
    OpenAI    & o4-mini-high                         & Yes       \\
    OpenAI    & gpt-5-minimal                        & Yes       \\
    OpenAI    & gpt-5-low                            & Yes       \\
    OpenAI    & gpt-5-medium                         & Yes       \\
    OpenAI    & gpt-5-high                           & Yes       \\
    xAI       & grok-4                               & Yes       \\
    \bottomrule
  \end{tabular}
\end{table}

\section{Experiment Resource Costs}
\label{costs}

Evaluations in this study were performed using commercial LLM APIs, including OpenAI gpt-4.1, o3, o4-mini-high, and the GPT-5 family (gpt-5-minimal, gpt-5-low, gpt-5-medium, gpt-5-high); Anthropic claude-3.7-sonnet and claude-3.7-sonnet\texttt{:}thinking; Google gemini-2.5-pro-preview-03-25 and gemini-2.5-flash-preview\texttt{:}thinking; and xAI grok-4. The total cost of API calls for model-based evaluation was approximately \$500.

\end{document}